\documentclass{ieeeaccess}
\usepackage[pdftex]{graphicx}
\usepackage{epsfig} 
\usepackage{amsthm}
\usepackage{amsmath}
\usepackage{amsfonts}
\usepackage{amssymb}
\usepackage{bm}
\usepackage{cases}
\usepackage{multirow}

\usepackage{multicol}
\usepackage{float,subfig}
\usepackage{caption}
\usepackage{cite}
\usepackage{url}
\usepackage{here}
\usepackage{stfloats} 
\usepackage{braket}
\usepackage[linesnumbered,ruled,vlined]{algorithm2e}
\usepackage{algorithmic}

\usepackage{multicol}
\usepackage[export]{adjustbox}

\usepackage{array}
\newcolumntype{C}[1]{>{\hfil}m{#1}<{\hfil}}

\usepackage{lscape} % figure/table rotation

\newcommand{\argmin}{\mathop{\rm \text{arg~min}}\limits}

\def\BibTeX{{\rm B\kern-.05em{\sc i\kern-.025em b}\kern-.08em
	T\kern-.1667em\lower.7ex\hbox{E}\kern-.125emX}}
\begin{document}
\history{Date of publication xxxx 00, 0000, date of current version xxxx 00, 0000.}
\doi{xxxx}

\title{Adaptive Resonance Theory-based Topological Clustering with a Divisive Hierarchical Structure Capable of Continual Learning}

\author{
	\uppercase{Naoki Masuyama}\authorrefmark{1}, \IEEEmembership{Member, IEEE},
	\uppercase{Narito Amako}\authorrefmark{2},
	\uppercase{Yuna Yamada}\authorrefmark{2},
	\uppercase{Yusuke Nojima}\authorrefmark{1}, \IEEEmembership{Member, IEEE},
	and \uppercase{Hisao Ishibuchi}\authorrefmark{3}, \IEEEmembership{Fellow, IEEE}.
}
\address[1]{Graduate School of Informatics, Osaka Metropolitan University,
	1-1 Gakuen-cho Naka-ku, Sakai-Shi, Osaka 599-8531, Japan (e-mails: \{masuyama,  nojima\}@omu.ac.jp)}
\address[2]{Graduate School of Engineering, Osaka Prefecture University,
	1-1 Gakuen-cho Naka-ku, Sakai-Shi, Osaka 599-8531, Japan (e-mails: \{narito.amako@ci.cs.osakafu-u.ac.jp, sbb01194@st.osakafu-u.ac.jp)}
\address[3]{Guangdong Provincial Key Laboratory of Brain-inspired Intelligent Computation, Department of Computer Science and Engineering, Southern University of Science and Technology, Shenzhen 518055, China (e-mail: hisao@sustech.edu.cn)}

\tfootnote{This research was supported by Ministry of Education, Culture, Sports, Science and Technology - JAPAN (MEXT) Leading Initiative for Excellent Young Researchers (LEADER). National Natural Science Foundation of China (Grant No. 61876075), Guangdong Provincial Key Laboratory (Grant No. 2020B121201001), the Program for Guangdong Introducing Innovative and Enterpreneurial Teams (Grant No. 2017ZT07X386), The Stable Support Plan Program of Shenzhen Natural Science Fund (Grant No. 20200925174447003), Shenzhen Science and Technology Program (Grant No. KQTD2016112514355531)}

\markboth
{Author \headeretal: Preparation of Papers for IEEE TRANSACTIONS and JOURNALS}
{Author \headeretal: Preparation of Papers for IEEE TRANSACTIONS and JOURNALS}

\corresp{Corresponding author: Hisao Ishibuchi (e-mail: hisao@sustech.edu.cn).}

\begin{abstract}
	Adaptive Resonance Theory (ART) is considered as an effective approach for realizing continual learning thanks to its ability to handle the plasticity-stability dilemma. In general, however, the clustering performance of ART-based algorithms strongly depends on the specification of a similarity threshold, i.e., a vigilance parameter, which is data-dependent and specified by hand. This paper proposes an ART-based topological clustering algorithm with a mechanism that automatically estimates a similarity threshold from the distribution of data points. In addition, for improving information extraction performance, a divisive hierarchical clustering algorithm capable of continual learning is proposed by introducing a hierarchical structure to the proposed algorithm. Experimental results demonstrate that the proposed algorithm has high clustering performance comparable with recently-proposed state-of-the-art hierarchical clustering algorithms.
\end{abstract}


\titlepgskip=-15pt

\maketitle

\section{Introduction}
\label{sec:introduction}
\PARstart{C}{lustering} is an essential technique to extract information from data. With the recent development of IoT technology, the importance of clustering increases as the availability of big data increases. Generally speaking, big data includes a wide range of useful information, e.g., hidden characteristics of data and explicit/implicit relationships among data points and/or attributes. To utilize useful information in big data with a pluralistic and appropriate information granularity, information extraction approaches such as hierarchical clustering are widely studied. In particular, since a divisive hierarchical clustering algorithm can adaptively generate partitions in response to the changes in the data distribution, the algorithm has the potential to realize continual learning, which is an essential ability to efficiently utilize big data.

Typical divisive hierarchical clustering algorithms with an adaptive cluster structure are Growing Hierarchical Self-Organizing Map (GHSOM) \cite{rauber02} and Growing Hierarchical Neural Gas (GHNG) \cite{palomo16b}. GHSOM uses Self-Organizing Map (SOM) \cite{kohonen82} as a base clustering approach while GHNG uses Growing Neural Gas (GNG) \cite{fritzke95}. Both GHSOM and GHNG can grow in the vertical and horizontal directions by constructing a tree-like structure based on the distribution of data points. However, in general, SOM-based and GNG-based clustering algorithms cannot avoid the plasticity-stability dilemma \cite{carpenter88}, i.e., the trade-off between catastrophic forgetting and continual learning of new information. 

In the research field of clustering, Adaptive Resonance Theory (ART) \cite{grossberg87} is a successful approach for handling the plasticity-stability dilemma. Among existing ART-based clustering algorithms \cite{carpenter91b, vigdor07, tscherepanow10, masuyama18, masuyama19b, masuyamaFTCA}, Fast Topological Correntropy-Induced Metric-based ART (FTCA) \cite{masuyamaFTCA} has superior clustering performance and functionality than the others. FTCA utilizes the Correntropy-Induced Metric (CIM) \cite{liu07} as a similarity measure for realizing a fast and stable self-organizing ability while maintaining an appropriate number of nodes. During a learning process, FTCA adaptively generates topological structures by using nodes and edges to efficiently extract information from a dataset.

In our previous study \cite{yamada20} inspired by GHNG \cite{palomo16b}, we have introduced Hierarchical FTCA (HFTCA) to improve the clustering performance of FTCA \cite{masuyamaFTCA}. Although HFTCA shows superior clustering performance than GHSOM and GHNG, HFTCA has a parameter (i.e., a similarity threshold) that significantly affects its clustering performance. In general, the optimal value of the similarity threshold depends on the dataset, which is a common difficulty in the use of ART-based clustering algorithms. Moreover, if an ART-based clustering algorithm has a hierarchical structure, it is necessary to specify a similarity threshold at each layer in advance. In this paper, to solve the above difficulty, we propose CIM-based ART with Edge and Age (CAEA), which has a mechanism to automatically estimate the similarity threshold from the distribution of data points. In addition, we also propose Hierarchical CAEA (HCAEA) for improving the clustering performance of CAEA. Thanks to the continual learning ability of CAEA, HCAEA is also capable of continual learning.

The contributions of this paper are summarized as follows:
\begin{itemize}
	\vspace{-1mm}
	\item[(i)] A new ART-based clustering algorithm, called CAEA, is proposed by introducing a mechanism to automatically estimate a similarity threshold from the distribution of data points.
	
	\item[(ii)] A new divisive hierarchical clustering algorithm, called HCAEA, is proposed by integrating CAEA into a hierarchical structure inspired by HFTCA.
	
	\item[(iii)] Empirical studies show that HCAEA has comparable clustering performance than recently-proposed state-of-the-art hierarchical algorithms: GHNG, HFTCA, and GH-EXIN.
\end{itemize}

The paper is organized as follows. Section \ref{sec:literature} presents the literature review for clustering algorithms and divisive hierarchical clustering algorithms. Section \ref{sec:proposedAlgorithm} describes details of the proposed algorithm and its hierarchical approach, namely CAEA and HCAEA. Section \ref{sec:experiment} presents simulation experiments to evaluate the information extraction performance of the proposed algorithms and the above-mentioned state-of-the-art algorithms by using real-world datasets. Section \ref{sec:discussion} summarizes characteristics of each algorithm based on the experimental results. Section \ref{sec:conclusion} concludes this paper.

\vspace{6mm}

\section{Literature Review}
\label{sec:literature}

\subsection{Clustering Algorithms Capable of Continual Learning}
\label{sec:literature_CL}
Clustering is one of the most widely used approaches for extracting information from data. Gaussian Mixture Model (GMM) \cite{mclachlan19} and $ k $-means \cite{lloyd82} are typical examples of clustering algorithms. Although GMM and $ k $-means are quite simple and highly applicable, they have a common limitation that the number of classes needs to be specified in advance. 

One effective approach for solving the drawback of GMM and $ k $-means is growing self-organizing clustering such as GNG \cite{fritzke95} and Self-Organizing Incremental Neural Network (SOINN) \cite{furao06}. GNG and SOINN can adaptively generate topological networks based on a distribution of data points. However, since these algorithms permanently insert new nodes into their networks for memorizing new information, they have the potential to forget learned information (i.e., catastrophic forgetting). This trade-off is called the plasticity-stability dilemma \cite{carpenter88}. 

Several clustering algorithms have been proposed to cope with the plasticity-stability dilemma. As a GNG-based algorithm, Grow When Required (GWR) \cite{marsland02} and gamma-GWR \cite{parisi17} are successful algorithms which appropriately calculate a similarity threshold to prevent an excessive node creation. As a SOINN-based algorithm, SOINN+ \cite{wiwatcharakoses20} and SOINN+ with ghost nodes (GSOINN+) \cite{wiwatcharakoses21} can detect clusters of arbitrary shapes in noisy data streams while avoiding catastrophic forgetting. Another successful approach is ART-based clustering algorithms such as Fuzzy ART \cite{carpenter91b}, Bayes ART \cite{vigdor07}, and their variants \cite{tscherepanow10, masuyama18}. In general, ART-based clustering algorithms show superior clustering performance than GNG-based and SOINN-based algorithms \cite{masuyama19b, masuyamaFTCA}. Moreover, because ART-based clustering algorithms can theoretically realize sequential and class-incremental learning without catastrophic forgetting, a number of ART-based clustering algorithms and their improvements have been proposed in both supervised learning \cite{tan14, matias18, matias21} and unsupervised learning \cite{carpenter91b, vigdor07, wang19, da20}. One common drawback of ART-based clustering algorithms that they need to specify a similarity threshold (i.e., a vigilance parameter). In most cases, the similarity threshold has a significant impact on their clustering performance while its optimal value depends on the dataset.

\subsection{Divisive Hierarchical Clustering Algorithms}
\label{sec:literature_HC}
There are two types of hierarchical approaches, i.e., an agglomerative approach and a divisive approach \cite{saxena17}. An agglomerative hierarchical clustering algorithm takes a bottom-up approach where data points are first considered as separate clusters and then merged into larger clusters based on the similarity between clusters. A divisive hierarchical clustering algorithm takes a top-down approach where all data points are considered as a cluster and then divided into smaller clusters based on the dissimilarity between clusters. In general, agglomerative hierarchical clustering algorithms require the entire dataset because of its learning procedure. Some well-known divisive hierarchical clustering algorithms such as the bisecting $ k $-means algorithm \cite{wang20} do not have this ability. That is, they need all data points in advance. Thus, they are not suitable for continual learning whereas they show high clustering performance. In contrast, divisive hierarchical clustering algorithms can be used for continual learning by adapting a cluster structure along with the change in the distribution of data points. Thanks to the above characteristics, divisive hierarchical clustering algorithms have the potential to realize continual learning.

In order to perform continual learning in a divisive hierarchical clustering algorithm, it is necessary to have the ability to adaptively extract information and generate a hierarchical structure in response to sequentially given data points. GHSOM \cite{rauber02} utilizes a growing SOM which can grows in the vertical and horizontal directions based on the distribution of data points. One drawback of GHSOM is raised by a SOM architecture, that is, it is difficult for GHSOM to represent multiple data distributions in a single SOM network. GHNG \cite{palomo16b} has successfully resolved the drawback of GHSOM by using GNG instead of SOM. One difficulty of GHNG is a large number of parameters. Growing Hierarchical EXcitatory+INhibitory (GH-EXIN) \cite{cirrincione20} requires only a few parameters while maintaining comparable performance to GHNG. In general, however, GNG-based clustering algorithms suffer from excessive node generation and high sensitivity to the presentation order of input data points. Moreover, the plasticity-stability dilemma \cite{carpenter88}, i.e., the trade-off between catastrophic forgetting and continual learning of new information, is another unavoidable problem. Applying an ART-based clustering algorithm as a basis of divisive hierarchical clustering is a promising approach because it can theoretically avoid the plasticity-stability dilemma. HFTCA \cite{yamada20} showed the superior clustering performance compared to GHSOM, GHNG, and GH-EXIN while maintaining a high information compression ratio. Although HFTCA is the state-of-the-art of ART-based divisive hierarchical clustering algorithm, HFTCA requires careful parameterization at each hierarchy in order to maintain good clustering performance.

\section{Preliminary Knowledge}
\label{sec:preliminary}
This section presents preliminary knowledge for a similarity measurement and a kernel density estimator that are used in the proposed algorithm.

\subsection{Correntropy and Correntropy-Induced Metric}
\label{sec:cimDefinition}
Correntropy \cite{liu07} provides a generalized similarity measure between two arbitrary data points $ \mathbf{x} = (x_{1},x_{2},\ldots,x_{d}) $ and $ \mathbf{y} = (y_{1},y_{2},\ldots,y_{d}) $ as follows:
\begin{equation}
	C(\mathbf{x}, \mathbf{y}) = \textbf{E} \left[ \kappa_{\sigma} \left( \mathbf{x}, \mathbf{y} \right) \right],
\end{equation}
where $ \textbf{E} \left[ \cdot \right] $ is the expectation operation, and $ \kappa_{\sigma} \left( \cdot \right) $ denotes a positive definite kernel with a bandwidth $ \sigma $. The correntropy can be estimated as follows:
\begin{equation}
	\hat{C}(\mathbf{x}, \mathbf{y}) = \frac{1}{d} \sum_{i=1}^{d} \kappa_{\sigma} \left( x_{i}, y_{i} \right).
	\label{eq:correntropy}
\end{equation}

In this paper, we use the following Gaussian kernel in the correntropy:
\begin{equation}
	\kappa_{\sigma} \left( x_{i}, y_{i} \right) = \exp \left[ - \frac{\left( x_{i} - y_{i} \right)^{2} }{2 \sigma^{2}} \right].
	\label{eq:gaussian}
\end{equation}

A nonlinear metric called CIM is derived from the correntropy \cite{liu07}. The CIM quantifies the similarity between two data points $ \mathbf{x} $ and $ \mathbf{y} $ as follows:
\begin{equation}
	\mathrm{CIM}\left(\mathbf{x}, \mathbf{y}, \sigma \right) = \left[ 1 - \hat{C}(\mathbf{x}, \mathbf{y}) \right]^{\frac{1}{2}}.
	\label{eq:defcim}
\end{equation}

Here, thanks to the Gaussian kernel without a coefficient $ \frac{1}{\sqrt{2\pi}\sigma} $ as defined in (\ref{eq:gaussian}), a range of the CIM is limited to $ \left[0,1 \right] $.

In general, the Euclidean distance suffers from the curse of dimensionality. However,  the CIM reduces this drawback thanks to the correntropy which calculates the similarity between two arbitrary data points by using a kernel function. Moreover, it has also been shown that the CIM with the Gaussian kernel has a high outlier rejection capability \cite{liu07}.

\subsection{Kernel Density Estimator}
\label{sec:kde}
In general, a bandwidth of a kernel function $ \sigma $ can be estimated based on a kernel density estimator \cite{henderson12}, which is defined as follows:
\begin{equation}
	\mathbf{\Sigma} = U(F_{\nu})  \mathbf{\Gamma} N^{-\frac{1}{2\nu+d}},
	\label{eq:sigmaEst1}
\end{equation}
\begin{equation}
	U(F_{\nu}) = \left( \frac{\pi ^{d/2} 2^{d+\nu-1}(\nu !)^{2}R(F)^{d}}{\nu \kappa_{\nu}^{2}(F)\left[(2\nu)!!+(d-1)(\nu!!)^{2}\right]} \right)^{\frac{1}{2\nu+d}},
	\label{eq:sigmaEst2}
\end{equation}
where $ \mathbf{\Gamma} $ denotes a rescale operator ($ d $-dimensional vector) which is defined by a standard deviation of attribute values of each attribute among $ N $ training data points. $ \nu $ is the order of the kernel. $ R(F) $ is a roughness function. $ \kappa_{\nu}(F) $ is the moment of the kernel. The details of the derivation of (\ref{eq:sigmaEst1}) can be found in \cite{henderson12}.

\begin{table*}[htbp]
	\centering
	\caption{Summary of notations}
	\renewcommand{\arraystretch}{1.2}
	\label{tab:notations}
	\scalebox{0.9}{
		\begin{tabular}{ll}
			\hline\hline
			Notation & Description \\
			\hline
			$ \mathcal{X} =$  $ \{ \mathbf{x}_{1}, \mathbf{x}_{2},\ldots, \mathbf{x}_{n}, \ldots \} $   & A set of training data points \\
			$ \mathbf{x}_{n} = $ $ \left( x_{n1}, x_{n2}, \ldots, x_{nd} \right) $  &  $ d $-dimensional training data point (the $ n $th data point)  \\ 
			$ \mathcal{Y} = \{\mathbf{y}_{1}, \mathbf{y}_{2}, \ldots, \mathbf{y}_{K}\} $  &  A set of prototype nodes \\ 
			$ \mathbf{y}_{k} = $ $ \left( y_{k1}, y_{k2}, \ldots, y_{kd} \right) $  & $ d $-dimensional prototype node (the $ k $th node) \\
			$ S = \{ \sigma_{1},\sigma_{2},\ldots, \sigma_{K}\} $  & A set of bandwidths for a kernel function \\ 
			$ \kappa_{\sigma} $  & Kernel function with a bandwidth $ \sigma $ \\ 
			\textrm{CIM}  &  Correntropy-Induced Metric\\ 
			$ k_{1} $, $ k_{2} $  & Indexes of the 1st and 2nd winner nodes  \\
			$ \mathbf{y}_{k_{1}} $, $ \mathbf{y}_{k_{2}} $ & The 1st and 2nd winner nodes \\
			$ V_{k_{1}} $, $ V_{k_{2}} $  & Similarities between a data point $ \mathbf{x}_{n} $ and winner nodes ($ \mathbf{y}_{k_{1}} $ and $ \mathbf{y}_{k_{2}} $) \\
			$ V_{\text{threshold}} $  &  Similarity threshold (a vigilance parameter) \\
			$ \mathcal{N}_{k_{1}} $ & A set of neighbor nodes of node $ \mathbf{y}_{k_{1}} $ \\
			$ M_{k_{1}} $  &  The number of data points that have accumulated by the node $ \mathbf{y}_{k_{1}} $ \\
			$ \lambda $ & Predefined interval for computing $ \sigma $ and deleting an isolated node \\
			$ a_{(k_{1}, k_{2})} $ & Age of edge between nodes $ \mathbf{y}_{k_{1}} $ and $ \mathbf{y}_{k_{2}} $ \\
			$ a_{\text{max}}$   & Predefined threshold of an age of edge \\
			%			$ e_{\left(k_{1}, k_{2}\right)} $ & Edge connection between nodes $ \mathbf{y}_{k_{1}} $ and $ \mathbf{y}_{k_{2}} $ \\
			$\mathbf{C} = \left(\mathbf{c}_{1}, \mathbf{c}_{2}, \ldots, \mathbf{c}_{K}\right) $ & Models of CAEA for constructing a hierarchical structure. \\
			\hline\hline
		\end{tabular}
	}
\end{table*}

\section{Proposed Algorithm}
\label{sec:proposedAlgorithm}
In this section, first CAEA is explained in detail. Next, a hierarchical approach of CAEA, namely HCAEA, is introduced. Table \ref{tab:notations} summarizes the main notations used in this paper.

\subsection{CIM-based ART with Edge and Age}
\label{sec:caea}
We use the following notations: A set of training data points is $ \mathcal{X} =$  $ \{ \mathbf{x}_{1}, \mathbf{x}_{2},\ldots, \mathbf{x}_{n}, \ldots \} $ where $ \mathbf{x}_{n} = $ $ ( x_{n1}, x_{n2}, \ldots,$  $ x_{nd} ) $ is a $ d $-dimensional feature vector. A set of prototype nodes in CAEA at the time of the presentation of a data point $ \mathbf{x}_{n} $ is $ \mathcal{Y} = \{\mathbf{y}_{1}, \mathbf{y}_{2},$  $ \ldots, \mathbf{y}_{K}\} $ ($ K \in \mathbb{Z}^{+} $) where a node $ \mathbf{y}_{k} = $ $ \left( y_{k1}, y_{k2}, \ldots, y_{kd} \right) $ has the same dimension as $ \mathbf{x}_{n} $. Furthermore, each node $ \mathbf{y}_{k} $ has an individual bandwidth $ \sigma $ for the CIM, i.e., $ S = \{ \sigma_{1},\sigma_{2},\ldots, \sigma_{K}\} $.

The learning procedure of CAEA is divided into four parts: 1) initialization process for nodes and a bandwidth of a kernel function in the CIM, 2) winner node selection, 3) vigilance test, and 4) node learning and edge construction. Each of them is explained in the following subsections.

\subsubsection{Initialization Process for Nodes and a Bandwidth of a Kernel Function in the CIM}
\label{sec:initProcess}
In the case that CAEA does not have any nodes, i.e., a set of prototype node $ \mathcal{Y} = \varnothing $, the 1st to $ \lambda/2 $th training data points $ X_{\text{init}} = \{ \mathbf{x}_{1}, \mathbf{x}_{2},\ldots \mathbf{x}_{\lambda/2} \} $ directly become prototype nodes, i.e., $ Y_{\text{init}} = \{\mathbf{y}_{1}, \mathbf{y}_{2}, \ldots, \mathbf{y}_{\lambda/2}\} $, where $ \lambda \in \mathbb{Z}^{+} $ is a predefined parameter of CAEA. Note that $ \lambda $ must be greater than or equal to 2, and if $ \lambda/2 $ is not an integer, it is rounded to the nearest integer. This parameter is also used for a node deletion process.

In an ART-based clustering algorithm, a vigilance parameter (i.e., a similarity threshold) plays an important role in a self-organizing process. Typically, the similarity threshold is data-dependent and specified by hand. On the other hand, CAEA uses the minimum pairwise CIM value between each of nodes in $ Y_{\text{init}} = \{\mathbf{y}_{1}, \mathbf{y}_{2}, \ldots, \mathbf{y}_{\lambda/2}\} $, and the average of pairwise CIM values is used as the similarity threshold $ V_{\text{threshold}} $, i.e.,
\begin{equation}
	V_{\text{threshold}} = \frac{1}{\lambda/2} \sum_{i=1}^{\lambda/2} \min_{j \neq i} \left[\mathrm{CIM}\left({\mathbf{y}}_{i}, {\mathbf{y}_{j}}, \sigma) \right)\right],
	\label{eq:pairwiseCIM}
\end{equation}
where $ \sigma $ is a bandwidth for a kernel function in the CIM.

%In general, a bandwidth of a kernel function $ \sigma $ can be estimated based on a kernel density estimator \cite{henderson12}, which is defined as follows:
%\begin{equation}
%	\mathbf{\Sigma} = U(F_{\nu})  \mathbf{\Gamma} N^{-\frac{1}{2\nu+d}},
%	\label{eq:sigmaEst1}
%\end{equation}
%\begin{equation}
%	U(F_{\nu}) = \left( \frac{\pi ^{d/2} 2^{d+\nu-1}(\nu !)^{2}R(F)^{d}}{\nu \kappa_{\nu}^{2}(F)\left[(2\nu)!!+(d-1)(\nu!!)^{2}\right]} \right)^{\frac{1}{2\nu+d}},
%	\label{eq:sigmaEst2}
%\end{equation}
%where $ \mathbf{\Gamma} $ denotes a rescale operator ($ d $-dimensional vector) which is defined by a standard deviation of attribute values of each attribute among $ N $ training data points. $ \nu $ is the order of the kernel. $ R(F) $ is a roughness function. $ \kappa_{\nu}(F) $ is the moment of the kernel. The details of the derivation of (\ref{eq:sigmaEst1}) can be found in \cite{henderson12}.

Here, $ \sigma $ is calculated by using (\ref{eq:sigmaEst1}) and (\ref{eq:sigmaEst2}). In this paper, we set $ N = \lambda/2 $, and we utilize the Gaussian kernel for the CIM. Therefore in (\ref{eq:sigmaEst2}), $ \nu = 2 $, $ R(F) = (2\sqrt{\pi})^{-1} $, and $ \kappa_{\nu}^{2}(F) = 1 $ are derived. As a result, we obtain:
\begin{equation}
	\mathbf{\Sigma} = \left( \frac{4}{2+d} \right)^{\frac{1}{4+d}}  \mathbf{\Gamma}  \left( \lambda/2 \right)^{-\frac{1}{4+d}}.
	\label{eq:SIGMA}
\end{equation}

In (\ref{eq:SIGMA}), $ \mathbf{\Sigma} $ contains the bandwidth of each attribute. In this paper, the median of $ \mathbf{\Sigma} $ is selected as a representative bandwidth of the Gaussian kernel in the CIM, i.e.,
\begin{equation}
	\sigma = \mathrm{median} \left( \mathbf{\Sigma} \right).
	\label{eq:sigma}
\end{equation}

In CAEA, the initial prototype nodes $ Y_{\text{init}} = \{\mathbf{y}_{1}, \mathbf{y}_{2}, \ldots,$  $ \mathbf{y}_{\lambda/2}\} $ have a common bandwidth of the Gaussian kernel in the CIM, i.e., $ S_{\text{init}} = \{ \sigma_{1},\sigma_{2},\ldots, \sigma_{\lambda/2}\} $. When a new node $ \mathbf{y}_{K+1} $ is generated from $ \mathbf{x}_{n} $, a bandwidth $ \sigma_{K+1} $ is estimated from the past $ \lambda/2 $ data points, i.e., $  \{ \mathbf{x}_{n-\lambda/2},\ldots, \mathbf{x}_{n-2}, \mathbf{x}_{n-1} \} $ by using (\ref{eq:sigmaEst1}) and (\ref{eq:sigmaEst2}). As a result, each new node has a different bandwidth $ \sigma $ depending on the distribution of training data points. Although the similarity threshold $ V_{\text{threshold}} $ depends on the distribution of the initial $ \lambda/2 $ training data points, we regard that an adaptive $ V_{\text{threshold}} $ estimation is realized by assigning a different bandwidth $ \sigma $, which affects the CIM value, for each node in response to the changes in the data distribution.

\subsubsection{Winner Node Selection}
\label{sec:winnerNS}
Once a data point $ \mathbf{x}_{n} $ is presented to CAEA with the prototype node set $ \mathcal{Y} = \{\mathbf{y}_{1}, \mathbf{y}_{2}, \ldots, \mathbf{y}_{K}\} $, two nodes which have a similar state to the data point $ \mathbf{x}_{n} $ are selected, namely, winner nodes $ \mathbf{y}_{k_{1}} $ and $ \mathbf{y}_{k_{2}} $. The winner nodes are determined based on the state of the CIM as follows:
\begin{equation}
	k_{1} = \argmin_{\mathbf{y}_{i} \in \mathcal{Y}}\left[ \mathrm{CIM}\left(\mathbf{x}_{n}, \mathbf{y}_{i}, \mathrm{mean}(S) \right) \right],
	\label{eq:winnerCIM1}
\end{equation}
\vspace{-5pt}
\begin{equation}
	\hspace{11.5mm} k_{2} = \argmin_{\mathbf{y}_{i} \in \mathcal{Y} \backslash \{\mathbf{y}_{k_{1}}\}}\left[ \mathrm{CIM}\left(\mathbf{x}_{n}, \mathbf{y}_{i}, \mathrm{mean}(S) \right) \right],
	\label{eq:winnerCIM2}
\end{equation}
\noindent where $ k_{1} $ and $ k_{2} $ denote the indexes of the 1st and 2nd winner nodes, i.e., $ \mathbf{y}_{k_{1}} $ and $ \mathbf{y}_{k_{2}} $, respectively. $ S $ is a bandwidth of the Gaussian kernel in the CIM for each node.

\subsubsection{Vigilance Test}
\label{sec:vigilanceTest}
Similarities between the data point $ \mathbf{x}_{n} $ and the 1st and 2nd winner nodes are defined as follows:
\begin{equation}
	V_{k_{1}} =  \mathrm{CIM}\left(\mathbf{x}_{n}, \mathbf{y}_{k_{1}}, \mathrm{mean}(S) \right),
	\label{eq:cim1}
\end{equation}
\vspace{-5pt}
\begin{equation}
	V_{k_{2}} = \mathrm{CIM}\left(\mathbf{x}_{n}, \mathbf{y}_{k_{2}}, \mathrm{mean}(S) \right).
	\label{eq:cim2}
\end{equation}

The vigilance test classifies the relationship between a data point and a node into three cases by using a predefined similarity threshold $ V_{\text{threshold}} $, i.e.,

\begin{itemize}
	[ 
	\setlength{\IEEElabelindent}{\dimexpr-\labelwidth-\labelsep}% Wrapping of text beyond first line of \item 
	\setlength{\itemindent}{\dimexpr\labelwidth+\labelsep}% identation for each new \item 
	\setlength{\listparindent}{\parindent}% Restore regular paragraph indentation 
	] 
	
	\item Case I \\
	\indent The similarity between the data point $ \mathbf{x}_{n} $ and the 1st winner node $ \mathbf{y}_{k_{1}} $ is larger (i.e., less similar) than $ V_{\text{threshold}} $, namely:
	\begin{equation}
		V_{\text{threshold}} < V_{k_{1}} < V_{k_{2}}.
		\label{eq:case1}
	\end{equation}
	
	\vspace{0.8mm}
	
	\item Case I\hspace{-.1em}I \\
	\indent The similarity between the data point $ \mathbf{x}_{n} $ and the 1st winner node $ \mathbf{y}_{k_{1}} $ is smaller (i.e., more similar) than $ V_{\text{threshold}} $, and the similarity between the data point $ \mathbf{x}_{n} $ and the 2nd winner node $ \mathbf{y}_{k_{2}} $ is larger (i.e., less similar) than $ V_{\text{threshold}} $, namely:
	\begin{equation}
		V_{k_{1}} \leq V_{\text{threshold}} < V_{k_{2}}.
		\label{eq:case2}
	\end{equation}
	
	\vspace{0.8mm}
	
	\item Case I\hspace{-.1em}I\hspace{-.1em}I \\
	\indent The similarities between the data point $ \mathbf{x}_{n} $ and the 1st and 2nd winner nodes are both smaller (i.e., more similar) than $ V_{\text{threshold}} $, namely:
	\begin{equation}
		V_{k_{1}} \leq V_{k_{2}} \leq V_{\text{threshold}}.
		\label{eq:case3}
	\end{equation}
	
\end{itemize}

\subsubsection{Node Learning and Edge Construction}
\label{sec:nodeLearning}

Depending on the result of the vigilance test, a different operation is performed.

If $ \mathbf{x}_{n} $ is classified as Case I by the vigilance test (i.e., (\ref{eq:case1}) is satisfied), a new node $ \mathbf{y}_{K+1} = \mathbf{x}_{n} $ is added to the prototype node set $ \mathcal{Y} = \{\mathbf{y}_{1}, \mathbf{y}_{2}, \ldots, \mathbf{y}_{K}\} $. A bandwidth $ \sigma_{K+1} $ for node $ \mathbf{y}_{K+1} $ is calculated by (\ref{eq:sigma}). In addition, the number of data points that have been accumulated by the node $ \mathbf{y}_{K+1} $ is initialized as $ M_{K+1} = 1 $.

If $ \mathbf{x}_{n} $ is classified as Case I\hspace{-1pt}I by the vigilance test (i.e., (\ref{eq:case2}) is satisfied), first, the age of each edge connected to the first winner node $ \mathbf{y}_{k_{1}} $ is updated as follows:
\begin{equation}
	a_{\left(k_{1}, j\right)} \gets a_{\left(k_{1}, j\right)} + 1 \quad \left(\forall j \in \mathcal{N}_{k_{1}}\right),
	\label{eq:edd_age}
\end{equation}
where $ \mathcal{N}_{k_{1}} $ is a set of nodes that are connected to $ \mathbf{y}_{k_{1}} $ by an edge. After updating the age of each of those edges, an edge whose age is greater than a predefined threshold $ a_{\text{max}} $ is deleted. Then, $ \mathbf{y}_{k_{1}} $ is updated as follows:
\begin{equation}
	\mathbf{y}_{k_{1}} \leftarrow \mathbf{y}_{k_{1}} + \frac{1}{M_{k_{1}}} \left( \mathbf{x}
	_{n} - \mathbf{y}_{k_{1}} \right).
	\label{eq:updateNodeWeight1}
\end{equation}

When updating the node, the difference between $ \mathbf{x}_{n} $ and $ \mathbf{y}_{n} $ is divided by $ M_{k_{1}} $. Thus, the larger $ M_{k_{1}} $ becomes, the smaller the node position changes. This is based on the idea that the information around a node, where data points are frequently given, is important and should be held by the node.

In Case I\hspace{-1pt}I, a counter $ M $ for the number of data points that have been accumulated by $ \mathbf{y}_{k_{1}} $ is also updated as follows:
\begin{equation}
	M_{k_{1}} \leftarrow M_{k_{1}} + 1.
	\label{eq:countNode}
\end{equation}

If $ {\mathbf{x}}_{n} $ is classified as Case I\hspace{-1pt}I\hspace{-1pt}I by the vigilance test (i.e., (\ref{eq:case3}) is satisfied), the same operations as Case I\hspace{-1pt}I (i.e., (\ref{eq:edd_age}), (\ref{eq:updateNodeWeight1}), and (\ref{eq:countNode})) are performed. In addition, the neighbor nodes of $ \mathbf{y}_{k_{1}} $ are updated as follows:
\begin{equation}
	{\mathbf{y}}_{j} \gets {\mathbf{y}}_{j} + \frac{1}{10 M_{j}} \left({\mathbf{x}}_{n}-{\mathbf{y}}_{j}\right) \quad \left(\forall j \in \mathcal{N}_{k_{1}}\right).
	\label{eq:updateNodeWeight2}
\end{equation}

Equation (\ref{eq:updateNodeWeight2}) has the same concept as (\ref{eq:updateNodeWeight1}), but it should be less affected by the data point than $ \mathbf{y}_{k_{1}} $ because it is the neighbor node of $ \mathbf{y}_{k_{1}} $. Thus, the value is multiplied by $ 1/10 $.

In Case I\hspace{-1pt}I\hspace{-1pt}I, moreover, if there is no edge between $ \mathbf{y}_{k_{1}} $ and $ \mathbf{y}_{k_{2}} $, a new edge is generated and its age is initialized as follows:
\begin{equation}
	a_{(k_{1}, k_{2})} \leftarrow 0.
	\label{eq:ageInit}
\end{equation} 

In the case that there is an edge between nodes $ \mathbf{y}_{k_{1}} $ and $ \mathbf{y}_{k_{2}} $, its age is also reset by (\ref{eq:ageInit}).

Apart from the above operations in Cases I-I\hspace{-1pt}I\hspace{-1pt}I, as a noise reduction purpose, the nodes without edges are deleted every time $ \lambda $ training data points are given.

The learning procedure of CAEA is summarized in Algorithm \ref{al:trainCAEA}. Note that the prediction procedure of CAEA is that an unknown data point is assigned to the class of its nearest neighbor node.

% CAEA Algorithm------------------------------------------------
\begin{algorithm}[htbp]
	\small
	\DontPrintSemicolon
	\caption{Learning Algorithm of CAEA}
	\label{al:trainCAEA}
	\KwIn{\\
		a set of training data points: $ \mathcal{X} =$  $ \{ \mathbf{x}_{1}, \mathbf{x}_{2},\ldots, \mathbf{x}_{n}, \ldots \} $ where $ \mathbf{x}_{n} = $ $ \left( x_{n1}, x_{n2}, \ldots, x_{nd} \right) $ $ (\bm{x}_{l} \in \Re^{d})$, \\
		%the current level of layer: $h$, \\
		the interval for computing $ \sigma $ and deleting an isolated node: $\lambda$, \\
		and the threshold of an age of edge: $a_{\text{max}}$.
	}
	\KwOut{\\
		the CAEA model.
		\begin{quote}
			\textbf{model contents} \\
			a set of generated nodes: $\mathcal{Y} = \{ \mathbf{y}_{1}, \mathbf{y}_{2}, \ldots, \mathbf{y}_{K} \} $ $ \left( K \in \mathbb{Z}^{+} \right) $, \\
			a set of bandwidths for a kernel function: $ S = \{ \sigma_{1},\sigma_{2},\ldots, \sigma_{K}\} $ \\ 
			a set of counters: $ M = \{ M_{1}, M_{2}, \ldots, M_{K} \} $, \\
			the matrix of edge connections: $\mathbf{e}$, \\
			and the matrix of edge age: $\mathbf{a}$.
		\end{quote}
	}
	\vspace{2mm}
	\SetKwBlock{Begin}{function}{end function}
	\Begin( \text{LearningCAEA($\mathcal{X}$, $\lambda$, $a_{\text{max}}$)}){
		\ForAll{ $ l \in {1, 2, \ldots, L} $ }{
			
			\uIf{ $ K < \lambda/2 $ }{
				Create the new node as $ {\mathbf{y}}_{K+1} = {\mathbf{x}}_{l} $. \\
				Calculate the bandwidth for a kernel function $ \sigma_{K+1} $ by (\ref{eq:SIGMA}) and (\ref{eq:sigma}). \\
				\If{ $ K = \lambda/2 $ }{
					Calculate the vigilance parameter $ V_\text{threshold} $ by (\ref{eq:pairwiseCIM}). \\
				}
			}\Else{
				Search the indexes of winner nodes $ k_{1} $ and $ k_{2} $ by (\ref{eq:winnerCIM1}) and (\ref{eq:winnerCIM2}), respectively. \\
				Update the edge age $a_{\left(k_{1}, j\right)}$ by (\ref{eq:edd_age}). \\
				\If{ $a_{\left(k_{1}, j\right)} > a_{\text{max}}$ }{
					Delete the edge. \\
				}
				\uIf{ \upshape{$ V_{k_{1}} > V_\text{threshold} $} }{
					Create the new node as $ {\mathbf{y}}_{K+1} = {\bm{x}}_{l} $.\\
					Calculate the bandwidth for a kernel function $ \sigma_{k+1} $ by (\ref{eq:SIGMA}) and (\ref{eq:sigma}). \\
				}\Else{
					Update the state of $ {\bm{y}}_{k_{1}} $ by (\ref{eq:updateNodeWeight1}). \\
					\If{ \upshape{$ V_{k_{2}} \leq V_\text{threshold} $} }{
						Update the state of neighbor nodes $ {\mathbf{y}}_{j} $ by (\ref{eq:updateNodeWeight2}). \\
						Create a new edge $e_{\left(k_{1}, k_{2}\right)}$ between $ {\mathbf{y}}_{k_{1}} $ and $ {\mathbf{y}}_{k_{2}} $. \\
					}
				}
			}
			%% start Topology Construction ---------------
			\If{ \upshape{the number of data point inputs $ l $ is multiple of a topology adjustment cycle $ \lambda $} }{
				\ForAll{ $ k \in {1, 2, \ldots, K} $ }{
					\If{ \upshape{$ {\mathbf{y}}_{k} $ does not have any edge} }{
						Remove $ {\mathbf{y}}_{k} $ from $ \mathcal{Y} $.
					}
				}
			}
		} %\ForALL
		\Return{ \upshape{the CAEA model} }.
	} %\SetKwBlock
\end{algorithm}
%\vspace{-5mm}

\subsection{Hierarchical Approach for CAEA}
\label{sec:hcaea}
The procedure for creating the hierarchical structure of HCAEA is as follows. First, CAEA is trained with a set of training data points $ \mathcal{X} =$  $ \{ \mathbf{x}_{1}, \mathbf{x}_{2},\ldots, \mathbf{x}_{n} \} $ to generate a topological network (nodes and edges) in the first layer. Supposing that $ \mathcal{Y} = \{\mathbf{y}_{1}, \mathbf{y}_{2},$  $ \ldots, \mathbf{y}_{K}\} $ is generated. Here, we preserve training data points that affect each node $\mathbf{y}_k$ during training process. As a result, we can define a new set of $ K $ training data point sets $ \mathcal{X}' =$  $ \{ \mathcal{X}_{1}, \mathcal{X}_{2}, \ldots, \mathcal{X}_{K} \} $ for the second layer, where $ \mathcal{X}' $ satisfies $ \mathcal{X}_{k} \neq \varnothing $, $ \bigcup_{\mathcal{X}_{k} \in \mathcal{X}'} \mathcal{X}_{k} = \mathcal{X} $, and $ \forall A, B \in \mathcal{X}' A \neq B  \implies A \cap B = \varnothing $. In the second layer, CAEA is independently trained by using each subset $ \mathcal{X}_{k} $ $ \left( k=1,2,\ldots,K \right) $ of training data points. This mechanism is used in a hierarchical manner for the training of CAEA. That is, CAEA in the ($h$+1)th layer is independently trained by using subsets of training data points, each of which is defined by the corresponding node in the $ h $th layer. The above procedure is repeated until an additional layer is no longer created by CAEA. That is, when the number of nodes (i.e., $K$) becomes two in the $h$th layer, it is too small to represent the distribution of data points. As a result, the training for creating the ($h$+1)th layer is not performed.

The learning procedure of HCAEA is summarized in Algorithm \ref{al:trainHCAEA}. Note that the prediction procedure of HCAEA is similar to CAEA. One difference from CAEA is that an unknown data point is assigned to the class of its nearest neighbor node which does not have a child node in the generated tree-like structure.

% HCAEA Algorithm------------------------------------------------
\begin{algorithm}[htbp]
	\small
	\DontPrintSemicolon
	\caption{Learning Algorithm of HCAEA}
	\label{al:trainHCAEA}
	\KwIn{\\
		a set of training data points: $ \mathcal{X} =$ $ \{ \mathbf{x}_{1}, \mathbf{x}_{2},\ldots, \mathbf{x}_{n}, \ldots \} $ where $ \mathbf{x}_{n} = $ $ \left( x_{n1}, x_{n2}, \ldots, x_{nd} \right) $ $ (\bm{x}_{l} \in \Re^{d})$, \\
		%the current level of layer: $h$, \\
		the interval for computing $\sigma$ and deleting an isolated node: $\lambda$, \\
		and the threshold of an age of edge: $a_{\text{max}}$. \\
	}
	\KwOut{\\
		the HCAEA model.
		\begin{quote}
			\textbf{model contents} \\
%			a set of training data points for the next layer: $ \mathcal{X}' =$  $ \{ \mathcal{X}_{1}, \mathcal{X}_{2}, \ldots, \mathcal{X}_{K} \} $ $ (\mathcal{X}' \subset \mathcal{X}) $, \\
			a set of training data points for the next layer: $ \mathcal{X}' =$  $ \{ \mathcal{X}_{1}, \mathcal{X}_{2}, \ldots, \mathcal{X}_{K} \} $ ($ \mathcal{X}_{k} \neq \varnothing $, $ \bigcup_{\mathcal{X}_{k} \in \mathcal{X}'} \mathcal{X}_{k} = \mathcal{X} $, and $ \forall A, B \in \mathcal{X}' A \neq B \implies A \cap B = \varnothing $), \\
			a set of generated nodes: $\mathcal{Y} = \{ {\mathbf{y}}_{1}, {\mathbf{y}}_{2}, \ldots, {\mathbf{y}}_{K} \} \left( K \in \mathbb{Z}^{+} \right)$, \\
			a set of bandwidths for a kernel function: $ S = \{ \sigma_{1},\sigma_{2},\ldots, \sigma_{K}\} $ \\ 
			a set of counters: $ M = \{ M_{1}, M_{2}, \ldots, M_{K} \} $, \\
			the matrix of edge connections: $\mathbf{e}$, \\
			the matrix of age of edge: $\mathbf{a}$, \\
			and a set of child models: $C = \{\mathbf{c}_{1}, \mathbf{c}_{2}, \ldots, \mathbf{c}_{K}\} $. \\
		\end{quote}
	}
	\vspace{2mm}
	\SetKwBlock{Begin}{function}{end function}
	\Begin( \text{LearningHCAEA($\mathcal{X}$, $\lambda$, $a_{\text{max}}$)}){
		\text{HCAEA model} = \textrm{LearningCAEA}($ \mathcal{X} $, $ \lambda $,  $ a_{\text{max}} $). \\
		\uIf{ $K \ge 2$ }{
			\ForAll{ $ k \in {1, 2, \ldots, K} $ }{
				Extract a set of data points $ \mathcal{X}_{k} $ that have accumulated by the node $ y_{k} $ as the training data points for the next layer. \\
				$ \mathbf{c}_{k} = $ \text{LearningHCAEA}($ \mathcal{X}_{k} $, $ \lambda $,  $ a_{\text{max}} $). \\
			}
			Update \text{the HCAEA model}. \\
		}\Else{
			\Return{}.
		}
		\Return{ \upshape{the HCAEA model} }. 
	}
\end{algorithm}
\vspace{-3mm}

\section{Simulation Experiments}
\label{sec:experiment}
This section presents quantitative comparisons focusing on the information extraction performance of CAEA, HCAEA, GHNG \cite{palomo16b}, GH-EXIN \cite{cirrincione20}, and HFTCA \cite{yamada20} based on the classification performance. In general, the evaluation of the clustering performance is subjective if a dataset does not have label information (i.e., each pattern in the dataset has no class label). In this paper, therefore, we use datasets with label information and perform classification tasks by using a clustering result as a classifier. This approach allows us to indirectly evaluate the clustering performance, that is, the performance of approximating the data distribution.

The source code of GHNG \footnote{\url{http://www.lcc.uma.es/~ejpalomo/software.html}}, GH-EXIN \footnote{\url{https://github.com/pietrobarbiero/ghexin}}, and HFTCA \footnote{\url{https://github.com/Masuyama-lab/HFTCA}} are provided by the authors of the related papers.

\subsection{Datasets}
\label{sec:datasets}
We utilize five synthetic datasets and nine real-world datasets selected from the commonly used clustering benchmarks \cite{franti18} and public repositories \cite{liu17, dua19}. Table \ref{tab:datasets} summarizes statistics of the datasets.

\begin{table}[htbp]
	\centering
	\renewcommand{\arraystretch}{1.2}
	\caption{
		Statistics of datasets for classification tasks
	}
	\label{tab:datasets}
	\scalebox{0.93}{ 
		\begin{tabular}{llrrr}
			\hline\hline
			\multirow{2}{*}{Type} & \multirow{2}{*}{Dataset} & Number of & Number of  & Number of \\
			& & Attributes & Classes & Instances\\
			\hline
			Synthetic & Aggretation & 2 & 7 & 788 \\
			& Compound & 2 & 6 & 299 \\
			& Hard Distribution & 2 & 3 & 1,500 \\
			& Jain & 2 & 2 & 373 \\
			& Pathbased & 2 & 3 & 300 \\
			\hline
			Real-world 
			& Breast Cancer & 30 & 2 & 569\\
			& COIL20 & 1,024 & 20 & 1,440\\
			& Iris & 4 & 3 & 150\\
			& Isolet & 617 & 26 & 1,560\\
			& OptDigits & 64 & 10 & 5,620\\
			& Seeds & 7 & 3 & 210\\
			& Sonar & 60 & 2 & 208\\
			& TOX171 & 5,748 & 4 & 171\\
			& Wine & 13 & 3 & 178 \\
			\hline\hline
		\end{tabular}
	}
\end{table}

\begin{table*}[htbp]
	\vspace{3mm}
	\caption{Parameter settings of each algorithm for classification tasks}
	\label{tab:paramAlgorithms}
	\footnotesize
	\centering
	\renewcommand{\arraystretch}{1.1}
	\scalebox{0.93}{
		\begin{tabular}{lllll}
			\hline\hline
			Algorithm & Parameter & Value & Grid Range & Description \\
			\hline
			GHNG & $\tau$ & grid search & \{0.001, 0.01, 0.05, 0.1, 0.15, 0.2, 0.25, 0.3, 0.35, 0.4, 0.45, 0.5\} & a learning coefficient \\
			& $\epsilon_{W}$ & 0.2 & --- & a learning coefficient \\
			& $\epsilon_{N}$ & 0.006 & --- & a learning coefficient \\
			& $\alpha$ & 0.5 & --- & a learning coefficient \\
			& $\beta$ & 0.995 & --- & a learning coefficient \\
			& ${\text{age}}_{\max}$ & 50 & --- & a maximum age of edge \\
			& $\lambda$ & 100 & --- & a node insertion cycle \\
			\hline
			GH-EXIN & $H_{\max}$ & grid search & \{0.00001, 0.0001, 0.001, 0.01, 0.1, 0.2, 0.3, 0.4, 0.5, 0.6, 0.7, 0.8, 0.9, 1.0\} & a task-dependent index \\
			& $\min_{\text{card}}$ & 100 & --- & a data cardinality \\
			& $H_{\text{perc}}$ & 0.5 & --- & a stopping criterion \\
			& $\epsilon_{W}$ & 0.2 & --- & a learning coefficient \\
			& $\epsilon_{N}$ & 0.006 & --- & a learning coefficient \\
			& ${\text{age}}_{\max}$ & 50 & --- & a maximum age of edge. \\
			\hline
			HFTCA & $V_{1}, V_{2}, \ldots, V_{L}$ & grid search & \{0.05, 0.1, 0.2, 0.3, 0.4, 0.5, 0.6, 0.7, 0.8, 0.9, 0.95\} & a vigilance parameter of each layer \\
			& $\lambda$ & 100 & --- & a topology construction cycle \\
			\hline
			CAEA & $\lambda$ & grid search & \{10, 12, 14, 16, 18, 20, 22, 24, 26, 28, 30\} & an interval for adapting $ \sigma $ \\
			& $a_{\text{max}}$ & 10 & --- & a maximum age of edge \\
			\hline
			HCAEA & $\lambda$ & grid search &  \{10, 12, 14, 16, 18, 20, 22, 24, 26, 28, 30\} & an interval for adapting $ \sigma $ \\
			& $a_{\text{max}}$ & 10 & --- & a maximum age of edge \\
			\hline\hline
		\end{tabular}
	}
	%	\vspace{-2mm}
\end{table*}

\begin{table*}[htbp]
	\vspace{4mm}
	\renewcommand{\arraystretch}{1.2}
	\centering{
		\caption{Parameters specified by grid search for classification tasks}
		\label{tab:paramClassificationGrid}
		\scalebox{1.0}{ 
			\begin{tabular}{ll|c|c|c|c|c}
				\hline\hline
				\multirow{2}{*}{Type} & \multirow{2}{*}{Dataset} & GHNG & GH-EXIN & HFTCA & CAEA & HCAEA \\
				&  & $ \tau $ & $ H_{\max} $ & $ (V_{1}, V_{2}, V_{3}, V_{4}, V_{5}) $ & $ \lambda $ & $ \lambda $ \\
				\hline
				Synthetic & Aggregation & 0.200 & 0.30000 & (0.95, 0.95, 0.60, --, --) & 30 & 30 \\
				& Compound & 0.100 & 0.00010 & (0.95, 0.95, 0.10, --, --) & 26 & 30 \\
				& Hard Distribution & 0.010 & 0.00001 & (0.70, 0.40, 0.30, 0.20, 0.10) & 26 & 28 \\
				& Jain & 0.300 & 0.10000 & (0.90, 0.90, 0.80, 0.30, --) & 26 & 26 \\
				& Pathbased & 0.001 & 0.01000 & (0.90, 0.80, 0.70, --, --) & 28 & 28 \\
				\hline
				Real-world & Breast Cancer & 0.300 & 0.00001 & (0.70, 0.60, 0.20, --, --) & 26 & 26 \\
				& COIL20 & 0.010 & 0.00001 &(0.40, --, --, --, --) & 26 & 26 \\
				& Iris & 0.500 & 0.00001 & (0.80, 0.70, 0.50, --, --) & 28 & 28 \\
				& Ioslet & 0.150 & 0.00100 & (0.70, 0.60, --, --, --) & 30 & 30 \\
				& OptDigits & 0.001 & 0.40000 & (0.70, --, --, --, --) & 20 & 20 \\
				& Seeds & 0.100 & 0.00010 & (0.80, 0.60, 0.40, --, --) & 28 & 16 \\
				& Sonar & 0.250 & 0.00100 & (0.40, --, --, --, --) & 24 & 24 \\
				& TOX171 & 0.300 & 0.00100 & N/A & 26 & 26 \\
				& Wine & 0.150 & 0.20000 & (0.90, 0.80, 0.50, --, --) & 24 & 28 \\
				\hline\hline
			\end{tabular}
		}
	}
	\\
	\vspace{1mm}
	\footnotesize \raggedright \hspace{24mm}N/A indicates that an algorithm could not build a predictive model.
	
	\hspace{24mm}A symbol "--" in HFTCA means that a layer is not generated.
	\vspace{3mm}
\end{table*}

\subsection{Parameter Specifications}
\label{sec:palamSpec}
CAEA, HCAEA, GHNG \cite{palomo16b}, GH-EXIN \cite{cirrincione20}, and HFTCA \cite{yamada20} have their own parameters, which influence their clustering performance. This section presents parameter specifications of each algorithm in detail.

Table \ref{tab:paramAlgorithms} summarizes parameter settings of each algorithm. In each algorithm, one parameter which has a large impact on the clustering performance is specified by grid search while the other parameters are specified as follows: In GHNG, the parameter $ \tau $ is unique and rest of them are the same as GNG. Therefore, we use the commonly used GNG parameter settings in GHNG. In GH-EXIN, some parameters are the same as GNG while $ \text{min}_{\text{card}} $ and $ H_{\text{perc}} $ are unique ones. $ \text{min}_{\text{card}} $ and $ H_{\text{perc}} $ are specified in \cite{cirrincione20}. In HFTCA, a topology construction cycle $ \lambda $ is the same as in \cite{yamada20}.

During grid search in our experiments, the training data points in each dataset are presented to each algorithm only once without pre-processing. In addition, each training data point is input under a stationary environment, i.e., the training data points are randomly selected from the entire data. For each parameter specification, we repeat the evaluation 20 times (i.e., 2$ \times $10-fold cross validation) with a different random seed for obtaining consistent averaging results.

Table \ref{tab:paramClassificationGrid} summarizes parameters which are specified by grid search. Using the parameter specifications in Tables \ref{tab:paramAlgorithms} and \ref{tab:paramClassificationGrid}, each algorithm shows the highest Normalized Mutual Information (NMI) \cite{strehl02} score for each dataset under a stationary environment. Note that a symbol "--" in HFTCA means that the layer is not generated.

\subsection{Classification Tasks}
\label{sec:classification}

\subsubsection{Conditions}
\label{sec:condition}
In order to demonstrate the information extraction performance of CAEA and HCAEA, we conduct classification tasks not only in a stationary environment but also in a non-stationary environment. In the stationary environment, the training data points are randomly selected from the entire dataset. In the non-stationary environment, the training data points are randomly selected from a specific class in the dataset, and the class is shifted sequentially.

Similar to grid search, we repeat the evaluation 20 times with no pre-processed training data points and a different random seed. The same parameter specifications are used in the stationary and non-stationary environments for each algorithm as explained in Tables \ref{tab:paramAlgorithms} and \ref{tab:paramClassificationGrid}. The classification performance is evaluated by Accuracy, NMI \cite{strehl02}, Adjusted Rand Index (ARI) \cite{hubert85}, and macro-$ \text{F}_{1} $.

As a statistical analysis, the Friedman test and Nemenyi post-hoc analysis \cite{demvsar06} are utilized. The Friedman test is used to test the null hypothesis that all algorithms perform equally. If the null hypothesis is rejected, the Nemenyi post-hoc analysis is then conducted. The Nemenyi post-hoc analysis is used for all pairwise comparisons based on the ranks of results over all the evaluation metrics for all datasets. Here, the null hypothesis is rejected at the significance level of $ 0.05 $ both in the Friedman test and the Nemenyi post-hoc analysis. All computations are carried out on Matlab 2020a with 2.2GHz Xeon Gold 6238R processor and 768GB RAM.

\subsubsection{Stationary Environment}
\label{sec:stationary}
\begin{figure}[htbp]
	\centering
	\includegraphics[width=3.3in]{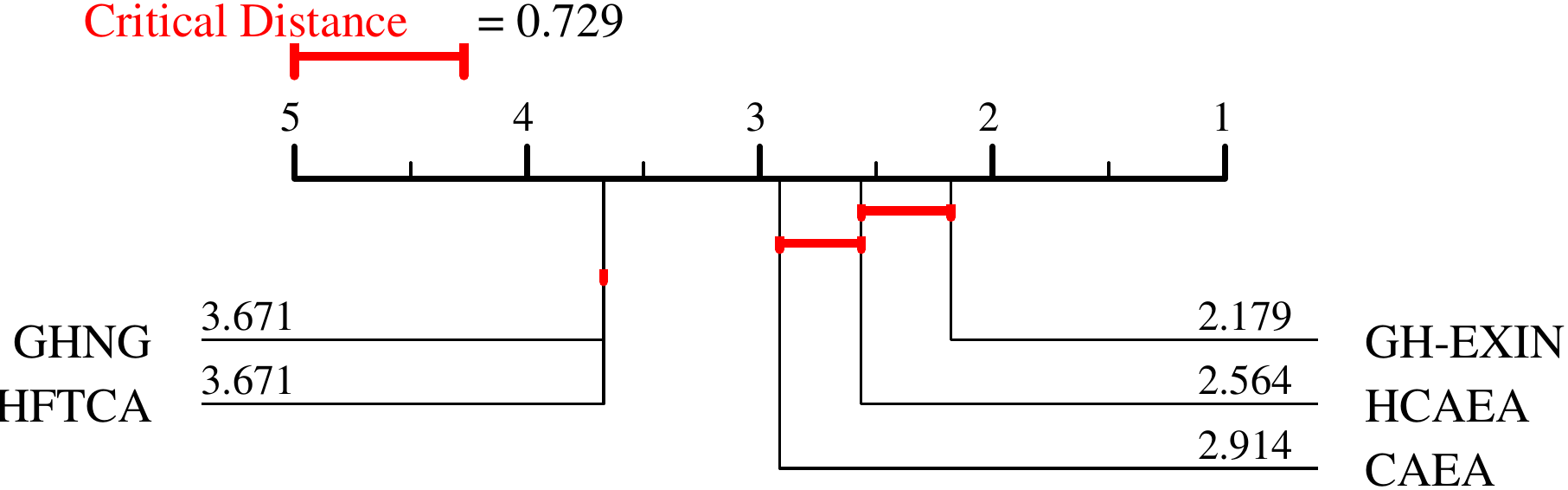}
	\caption{Critical difference diagram of classification tasks in the stationary environment.}
	\label{fig:CD_S}
	\vspace{-2mm}
\end{figure}

Table \ref{tab:ClassificationS} shows the results of the classification performance in the stationary environment. The best value in each metric is indicated by bold. The standard deviation is indicated in parentheses. N/A indicates that an algorithm could not build a predictive model. As an overall trend, GH-EXIN and HCAEA show better performance than CAEA, GHNG and HFTCA. Here, the null hypothesis is rejected on the Friedman test over all the evaluation metrics and datasets. Thus, we apply the Nemenyi post-hoc analysis. Fig. \ref{fig:CD_S} shows a critical difference diagram based on the classification performance including all the evaluation metrics and datasets. Better performance is shown by lower average ranks, i.e., on the right side of a critical distance diagram. In theory, different methods within a critical distance (i.e., a red line) do not have a statistically significance difference \cite{demvsar06}. In Fig. \ref{fig:CD_S}, GH-EXIN shows the best performance but there is no statistically significant difference from HCAEA. Comparing HCAEA and CAEA, although there is no statistically significant difference, HCAEA shows a lower rank. This suggests that a divisive hierarchical structure of HCAEA has a positive impact on the classification/clustering performance.

\begin{table*}[htbp]
	\renewcommand{\arraystretch}{1.15}
	\centering{
		\caption{
			Results of the classification performance in the stationary environment
		}
		\label{tab:ClassificationS}
		\scalebox{1.0}{ 
			\begin{tabular}{lll|c|c|c|c|c}
				\hline\hline
				Type & Dataset & Metric & GHNG & GH-EXIN & HFTCA & CAEA & HCAEA \\
				\hline
				Synthetic & Aggregation & Accuracy & 0.896 (0.048) & 0.993 (0.009) & 0.489 (0.059) & 0.957 (0.051) & \textbf{0.996 (0.006)} \\
				&  & NMI & 0.883 (0.047) & 0.987 (0.016) & 0.337 (0.049) & 0.948 (0.057) & \textbf{0.992 (0.012)} \\
				&  & ARI & 0.825 (0.074) & 0.986 (0.018) & 0.227 (0.063) & 0.929 (0.077) & \textbf{0.992 (0.013)} \\
				&  & macro-$ \text{F}_{1} $ & 0.701 (0.107) & 0.987 (0.018) & 0.300 (0.045) & 0.872 (0.116) & \textbf{0.992 (0.014)} \\
				\cline{2-8}
				& Compound & Accuracy & 0.783 (0.107) & 0.930 (0.042) & 0.574 (0.072) & 0.871 (0.055) & \textbf{0.956 (0.030)} \\
				&  & NMI & 0.796 (0.080) & 0.906 (0.044) & 0.443 (0.081) & 0.861 (0.058) & \textbf{0.936 (0.039)} \\
				&  & ARI & 0.704 (0.146) & 0.888 (0.070) & 0.361 (0.122) & 0.797 (0.088) & \textbf{0.916 (0.062)} \\
				&  & macro-$ \text{F}_{1} $ & 0.589 (0.185) & 0.911 (0.050) & 0.329 (0.055) & 0.794 (0.107) & \textbf{0.945 (0.049)} \\
				\cline{2-8}
				& HardDistribution & Accuracy & 0.989 (0.009) & \textbf{0.995 (0.006)} & 0.992 (0.007) & 0.987 (0.008) & 0.984 (0.012) \\
				&  & NMI & 0.955 (0.031) & \textbf{0.976 (0.025)} & 0.967 (0.030) & 0.950 (0.029) & 0.939 (0.039) \\
				&  & ARI & 0.968 (0.025) & \textbf{0.984 (0.017)} & 0.976 (0.022) & 0.962 (0.023) & 0.953 (0.034) \\
				&  & macro-$ \text{F}_{1} $ & 0.989 (0.009) & \textbf{0.995 (0.005)} & 0.992 (0.007) & 0.987 (0.008) & 0.984 (0.012) \\
				\cline{2-8}
				& Jain & Accuracy & 0.887 (0.068) & \textbf{1.000 (0.000)} & 0.842 (0.055) & 0.991 (0.016) & 0.999 (0.006) \\
				&  & NMI & 0.503 (0.224) & \textbf{1.000 (0.000)} & 0.304 (0.156) & 0.937 (0.103) & 0.991 (0.039) \\
				&  & ARI & 0.569 (0.230) & \textbf{1.000 (0.000)} & 0.417 (0.174) & 0.959 (0.069) & 0.994 (0.025) \\
				&  & macro-$ \text{F}_{1} $ & 0.835 (0.107) & \textbf{1.000 (0.000)} & 0.771 (0.087) & 0.986 (0.023) & 0.998 (0.007) \\
				\cline{2-8}
				& Pathbased & Accuracy & 0.653 (0.095) & 0.955 (0.038) & 0.548 (0.101) & 0.905 (0.071) & \textbf{0.982 (0.025)} \\
				&  & NMI & 0.476 (0.109) & 0.871 (0.105) & 0.181 (0.094) & 0.788 (0.114) & \textbf{0.947 (0.071)} \\
				&  & ARI & 0.363 (0.137) & 0.862 (0.113) & 0.124 (0.098) & 0.753 (0.156) & \textbf{0.943 (0.080)} \\
				&  & macro-$ \text{F}_{1} $ & 0.594 (0.105) & 0.956 (0.039) & 0.527 (0.105) & 0.897 (0.074) & \textbf{0.982 (0.024)} \\
				\hline
				Real-world & Breast Cancer & Accuracy & 0.889 (0.057) & \textbf{0.914 (0.033)} & 0.887 (0.044) & 0.910 (0.039) & 0.909 (0.041) \\
				&  & NMI & 0.541 (0.176) & \textbf{0.588 (0.118)} & 0.503 (0.161) & 0.588 (0.136) & 0.586 (0.139) \\
				&  & ARI & 0.606 (0.172) & \textbf{0.679 (0.110)} & 0.594 (0.141) & 0.667 (0.132) & 0.665 (0.135) \\
				&  & macro-$ \text{F}_{1} $ & 0.878 (0.058) & \textbf{0.904 (0.037)} & 0.875 (0.050) & 0.900 (0.043) & 0.900 (0.043) \\
				\cline{2-8}
				& COIL20 & Accuracy & 0.433 (0.056) & \textbf{0.830 (0.043)} & 0.705 (0.057) & 0.555 (0.062) & 0.555 (0.062) \\
				&  & NMI & 0.716 (0.037) & \textbf{0.892 (0.027)} & 0.832 (0.023) & 0.735 (0.035) & 0.735 (0.035) \\
				&  & ARI & 0.416 (0.060) & \textbf{0.765 (0.069)} & 0.627 (0.060) & 0.446 (0.072) & 0.446 (0.072) \\
				&  & macro-$ \text{F}_{1} $ & 0.333 (0.049) & \textbf{0.815 (0.045)} & 0.654 (0.056) & 0.462 (0.068) & 0.462 (0.068) \\
				\cline{2-8}
				& Iris & Accuracy & 0.883 (0.068) & 0.960 (0.055) & 0.947 (0.056) & \textbf{0.967 (0.046)} & 0.960 (0.050) \\
				&  & NMI & 0.806 (0.070) & 0.917 (0.111) & 0.888 (0.099) & \textbf{0.927 (0.096)} & 0.913 (0.106) \\
				&  & ARI & 0.747 (0.118) & 0.896 (0.143) & 0.873 (0.115) & \textbf{0.909 (0.128)} & 0.895 (0.134) \\
				&  & macro-$ \text{F}_{1} $ & 0.846 (0.112) & 0.950 (0.075) & 0.924 (0.100) & \textbf{0.960 (0.061)} & 0.950 (0.067) \\
				\cline{2-8}
				& Isolet & Accuracy & 0.505 (0.078) & \textbf{0.812 (0.014)} & 0.482 (0.031) & 0.332 (0.032) & 0.332 (0.032) \\
				&  & NMI & 0.681 (0.029) & \textbf{0.828 (0.011)} & 0.613 (0.023) & 0.515 (0.020) & 0.515 (0.020) \\
				&  & ARI & 0.423 (0.044) & \textbf{0.691 (0.021)} & 0.351 (0.026) & 0.199 (0.024) & 0.199 (0.024) \\
				&  & macro-$ \text{F}_{1} $ & 0.465 (0.090) & \textbf{0.808 (0.013)} & 0.444 (0.042) & 0.274 (0.034) & 0.274 (0.034) \\
				\cline{2-8}
				& OptDigits & Accuracy & 0.899 (0.016) & \textbf{0.953 (0.010)} & 0.760 (0.021) & 0.841 (0.085) & 0.835 (0.112) \\
				&  & NMI & 0.832 (0.026) & \textbf{0.912 (0.018)} & 0.637 (0.033) & 0.766 (0.077) & 0.757 (0.109) \\
				&  & ARI & 0.797 (0.031) & \textbf{0.901 (0.022)} & 0.565 (0.042) & 0.710 (0.114) & 0.701 (0.162) \\
				&  & macro-$ \text{F}_{1} $ & 0.898 (0.017) & \textbf{0.953 (0.010)} & 0.753 (0.018) & 0.832 (0.099) & 0.826 (0.125) \\
				\cline{2-8}
				& Seeds & Accuracy & \textbf{0.881 (0.059)} & 0.843 (0.064) & 0.864 (0.060) & 0.869 (0.065) & 0.876 (0.059) \\
				&  & NMI & \textbf{0.715 (0.123)} & 0.639 (0.132) & 0.691 (0.117) & 0.703 (0.136) & 0.702 (0.113) \\
				&  & ARI & \textbf{0.669 (0.149)} & 0.596 (0.146) & 0.654 (0.141) & 0.650 (0.163) & 0.665 (0.142) \\
				&  & macro-$ \text{F}_{1} $ & \textbf{0.872 (0.064)} & 0.825 (0.074) & 0.848 (0.070) & 0.862 (0.067) & 0.865 (0.060) \\
				\cline{2-8}
				& Sonar & Accuracy & 0.493 (0.115) & 0.664 (0.099) & 0.625 (0.089) & \textbf{0.688 (0.116)} & \textbf{0.688 (0.116)} \\
				&  & NMI & 0.038 (0.043) & 0.149 (0.120) & 0.083 (0.070) & \textbf{0.182 (0.162)} & \textbf{0.182 (0.162)} \\
				&  & ARI & 0.014 (0.034) & 0.111 (0.117) & 0.064 (0.076) & \textbf{0.160 (0.162)} & \textbf{0.160 (0.162)} \\
				&  & macro-$ \text{F}_{1} $ & 0.453 (0.123) & 0.651 (0.100) & 0.614 (0.087) & \textbf{0.674 (0.121)} & \textbf{0.674 (0.121)} \\
				\cline{2-8}
				& TOX171 & Accuracy & 0.346 (0.119) & 0.424 (0.086) & N/A & \textbf{0.471 (0.139)} & \textbf{0.471 (0.139)} \\
				&  & NMI & 0.239 (0.138) & 0.357 (0.110) & N/A & \textbf{0.359 (0.126)} & \textbf{0.359 (0.126)} \\
				&  & ARI & 0.080 (0.069) & \textbf{0.143 (0.137)} & N/A & 0.125 (0.113) & 0.125 (0.113) \\
				&  & macro-$ \text{F}_{1} $ & 0.250 (0.098) & 0.367 (0.092) & N/A & \textbf{0.387 (0.136)} & \textbf{0.387 (0.136)} \\
				\cline{2-8}
				& Wine & Accuracy & 0.657 (0.086) & 0.657 (0.098) & 0.874 (0.081) & \textbf{0.876 (0.089)} & 0.846 (0.122) \\
				&  & NMI & 0.475 (0.118) & 0.481 (0.125) & \textbf{0.729 (0.136)} & 0.720 (0.179) & 0.684 (0.191) \\
				&  & ARI & 0.334 (0.140) & 0.348 (0.167) & 0.644 (0.190) & \textbf{0.655 (0.218)} & 0.610 (0.248) \\
				&  & macro-$ \text{F}_{1} $ & 0.586 (0.081) & 0.629 (0.094) & \textbf{0.883 (0.076)} & 0.869 (0.097) & 0.835 (0.133) \\
				\hline\hline
			\end{tabular}
		}
	}
	%	\vspace{-4mm}
	\\
	\vspace{1mm}
	\footnotesize \raggedright \hspace{13mm}The best value in each metric is indicated by bold. The standard deviation is indicated in parentheses.
	
	\hspace{13mm}N/A indicates that the corresponding algorithm could not build a predictive model.
\end{table*}

Table \ref{tab:NodeS} shows the number of generated nodes by each algorithm in the stationary environment. As a general tendency, GH-EXIN generates a large number of nodes and GHNG generates a small number of nodes. The number of nodes in HCAEA is larger than that of CAEA because of its hierarchical structure. Note that the number of layers of HCAEA is automatically specified by the algorithm. In this experiment, HCAEA generates a single layer for COIL20, Isolet, Sonar, TOX171, and Wine datasets. These datasets have a small number of training data points compared to the number of attributes. From the above observations, it can be concluded that HCAEA can adaptively and automatically specify the sufficient number of nodes and layers for extracting information.

Table \ref{tab:TimeS} summarizes the results of training time in the stationary environment. In general, GHNG is faster than all the other algorithms because the number of generated nodes is smaller than the others. Regarding GH-EXIN, the computation speed depends on the number of attributes in the dataset. Namely, GH-EXIN is fast when the number of attributes is small while it is slow when the number of attributes is large. The computational efficiency of HCAEA and CAEA depends on the number of generated nodes. In short, the computation speed is slow when the number of generated nodes is large, and vice versa.

\begin{table*}[htbp]
	\centering
	\renewcommand{\arraystretch}{1.1}
	\caption{
		Results of the number of leaf nodes in the stationary environment
	}
	\label{tab:NodeS}
	\scalebox{1.0}{ 
		\begin{tabular}{ll|c|c|c|c|c}
			\hline\hline
			Type & Dataset & GHNG & GH-EXIN & HFTCA & CAEA & HCAEA \\
			\hline
			Synthetic & Aggregation & 23.2 (1.9) & 141.2 (9.4) & 108.8 (12.1) & 35.5 (4.1) & 445.7 (55.2) \\
			& Compound & 14.8 (1.8) & 136.5 (9.7) & 64.7 (5.7) & 28.0 (2.8) & 33.9 (15.5) \\
			& HardDistribution & 14.0 (0.0) & 294.7 (40.5) & 79.6 (10.4) & 30.1 (1.6) & 30.1 (1.6) \\
			& Jain & 11.9 (1.4) & 95.1 (6.9) & 50.1 (5.8) & 25.4 (3.4) & 245.1 (35.2) \\
			& Pathbased & 39.3 (3.2) & 311.9 (9.4) & 43.1 (5.1) & 33.2 (4.9) & 685.6 (73.9) \\
			\hline
			Real-world & Breast Cancer & 3.0 (0.0) & 33.8 (3.6) & 54.0 (6.1) & 25.7 (3.3) & 65.8 (15.0) \\
			& COIL20 & 65.5 (13.4) & 2255.8 (144.8) & 34.6 (3.5) & 32.4 (2.5) & 32.4 (2.5) \\
			& Iris & 9.8 (1.5) & 86.1 (5.1) & 25.9 (5.7) & 25.6 (3.9) & 218.7 (29.9) \\
			& Isolet & 154.7 (4.1) & 1226.9 (134.8) & 76.1 (14.3) & 38.3 (7.2) & 400.5 (127.1) \\
			& OptDigits & 8.5 (1.5) & 72.4 (6.9) & 723.9 (19.3) & 25.6 (2.6) & 168.7 (16.7) \\
			& Seeds & 3.6 (1.2) & 49.2 (4.6) & 45.2 (6.1) & 25.4 (2.8) & 69.4 (19.6) \\
			& Sonar & 3.8 (1.3) & 44.7 (8.1) & 15.7 (2.3) & 23.9 (2.2) & 23.9 (2.2) \\
			& TOX171 & 3.0 (0.0) & 31.0 (8.2) & N/A & 27.3 (3.2) & 27.3 (3.2) \\
			& Wine & 3.9 (1.4) & 41.6 (4.7) & 35.0 (3.5) & 21.9 (3.4) & 28.4 (8.8) \\
			\hline\hline
		\end{tabular}
	}
	\\
	\vspace{1mm}
	\footnotesize \raggedright \hspace{24mm}The standard deviation is indicated in parentheses.
	
	\hspace{24mm}N/A indicates that an algorithm could not build a predictive model.
\end{table*}

\begin{table*}[htbp]
	\centering
	\renewcommand{\arraystretch}{1.1}
	\caption{
		Results of training time on CPU [sec] in the stationary environment
	}
	\label{tab:TimeS}
	\scalebox{1.0}{ 
		\begin{tabular}{ll|c|c|c|c|c}
			\hline\hline
			Type & Dataset & GHNG & GH-EXIN & HFTCA & CAEA & HCAEA \\
			\hline
			Synthetic & Aggregation & 17.078 (1.408) & 1.112 (0.093) & 16.860 (1.656) & 7.736 (0.512) & 16.709 (2.077) \\
			& Compound & 0.709 (0.070) & 2.512 (0.199) & 3.306 (0.209) & 0.558 (0.064) & 0.571 (0.049) \\
			& HardDistribution & 0.124 (0.009) & 193.311 (13.438) & 42.863 (5.481) & 1.679 (0.101) & 1.703 (0.091) \\
			& Jain & 3.012 (0.263) & 0.824 (0.192) & 2.951 (0.212) & 1.624 (0.121) & 3.275 (0.248) \\
			& Pathbased & 60.086 (4.993) & 6.375 (1.199) & 2.109 (0.168) & 27.056 (2.211) & 71.756 (12.270) \\
			\hline
			Real-world & Breast Cancer & 0.241 (0.019) & 0.267 (0.023) & 0.657 (0.087) & 0.221 (0.023) & 0.309 (0.028) \\
			& COIL20 & 1.667 (0.187) & 500.936 (116.089) & 1.720 (0.100) & 11.124 (0.532) & 11.120 (0.518) \\
			& Iris & 2.761 (0.287) & 0.727 (0.066) & 0.287 (0.029) & 1.489 (0.194) & 3.415 (0.322) \\
			& Isolet & 34.098 (2.078) & 173.905 (66.630) & 9.656 (0.731) & 38.382 (4.096) & 60.154 (7.300) \\
			& OptDigits & 2.082 (0.203) & 0.804 (0.325) & 236.825 (13.707) & 1.293 (0.087) & 2.374 (0.135) \\
			& Seeds & 0.268 (0.034) & 0.432 (0.034) & 0.273 (0.019) & 0.227 (0.018) & 0.377 (0.114) \\
			& Sonar & 0.030 (0.003) & 1.011 (0.104) & 0.040 (0.002) & 0.055 (0.003) & 0.051 (0.002) \\
			& TOX171 & 0.002 (0.000) & 182.100 (29.984) & N/A & 0.024 (0.001) & 0.021 (0.002) \\
			& Wine & 0.105 (0.015) & 0.388 (0.037) & 0.108 (0.006) & 0.103 (0.031) & 0.096 (0.008) \\
			\hline\hline
		\end{tabular}
	}
	\\
	\vspace{1mm}
	\footnotesize \raggedright \hspace{13mm}The standard deviation is indicated in parentheses.
	
	\hspace{13mm}N/A indicates that an algorithm could not build a predictive model.
	\vspace{-2mm}
\end{table*}

\subsubsection{Non-stationary Environment}
\label{sec:nonstationary}
In the case of the non-stationary environment, a continual learning ability is required to maintain the classification/clustering performance because the training data points are randomly selected from a specific class of the dataset, and the class is shifted sequentially.

In general, whether an environment is stationary or non-stationary is unknown and dynamic. In other words, an algorithm should have characteristics that do not deteriorate its classification/clustering performance in any environment. For this reason, the parameter specifications in the non-stationary environment are the same as in the stationary environment, i.e., Tables \ref{tab:paramAlgorithms} and \ref{tab:paramClassificationGrid}.

\begin{figure}[t]
	\centering
	\includegraphics[width=3.3in]{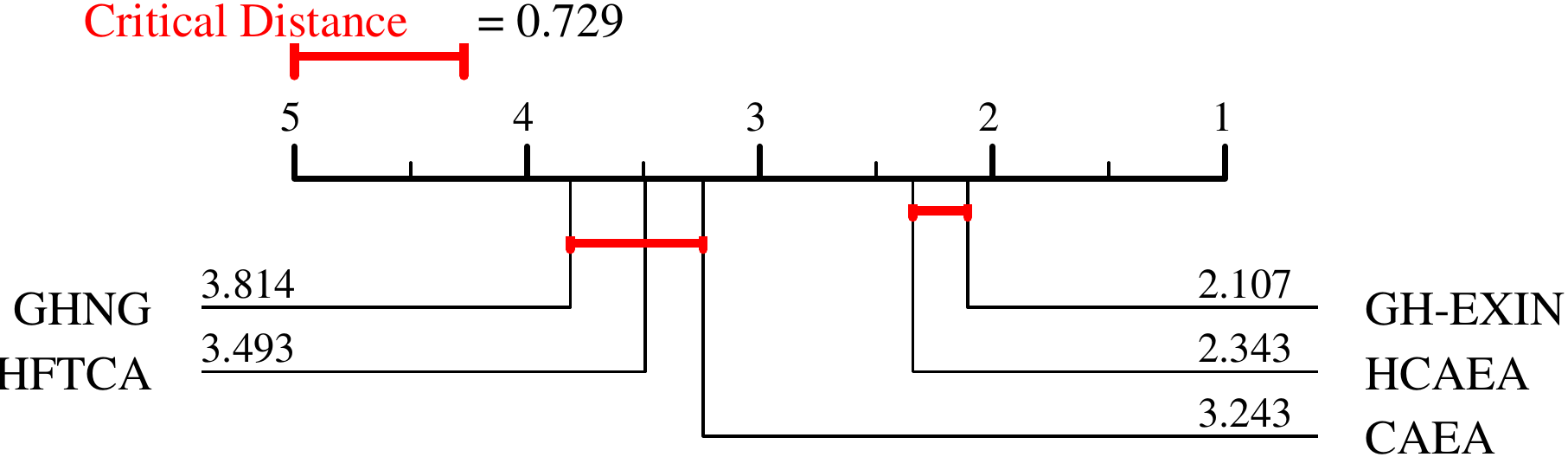}
	\caption{Critical difference diagram of classification tasks in the non-stationary environment.}
	\label{fig:CD_NS}
	\vspace{-2mm}
\end{figure}

Table \ref{tab:ClassificationNS} shows the results of the classification performance in the non-stationary environment. The best value in each metric is indicated by bold for each dataset. The standard deviation is indicated in parentheses. N/A indicates that an algorithm could not build a predictive model. Similar to the results in the stationary environment, the classification performance of GH-EXIN and HCAEA is generally better than CAEA, GHNG and HFTCA. Here, the null hypothesis is rejected on the Friedman test over all the evaluation metrics and datasets. Thus, we apply the Nemenyi post-hoc analysis. Fig. \ref{fig:CD_NS} shows a critical difference diagram based on the classification performance including all the evaluation metrics and datasets. Similar to the results in the stationary environment, GH-EXIN shows the best performance but there is no statistically significant difference from HCAEA. Regarding HCAEA and CAEA, there is a statistically significant difference between these two algorithms. This result indicates that HCAEA has the stable and superior information extraction performance than CAEA in the non-stationary environments.

\begin{table*}[htbp]
	\renewcommand{\arraystretch}{1.15}
	\centering{
		\caption{
			Results of the classification performance in the non-stationary environment
		}
		\label{tab:ClassificationNS}
		\scalebox{1.0}{ 
			\begin{tabular}{lll|c|c|c|c|c}
				\hline\hline
				Type & Dataset & Metric & GHNG & GH-EXIN & HFTCA & CAEA & HCAEA \\
				\hline
				Synthetic & Aggregation & Accuracy & 0.886 (0.043) & 0.992 (0.010) & 0.478 (0.073) & 0.979 (0.026) & \textbf{0.996 (0.006)} \\
				&  & NMI & 0.858 (0.045) & 0.986 (0.016) & 0.362 (0.043) & 0.964 (0.035) & \textbf{0.992 (0.012)} \\
				&  & ARI & 0.795 (0.065) & 0.984 (0.019) & 0.237 (0.062) & 0.956 (0.041) & \textbf{0.993 (0.012)} \\
				&  & macro-$ \text{F}_{1} $ & 0.693 (0.100) & 0.984 (0.034) & 0.345 (0.072) & 0.962 (0.059) & \textbf{0.993 (0.013)} \\
				\cline{2-8}
				& Compound & Accuracy & 0.781 (0.094) & 0.949 (0.035) & 0.560 (0.061) & 0.936 (0.060) & \textbf{0.959 (0.035)} \\
				&  & NMI & 0.791 (0.086) & 0.919 (0.048) & 0.430 (0.079) & 0.909 (0.070) & \textbf{0.935 (0.050)} \\
				&  & ARI & 0.701 (0.131) & 0.905 (0.068) & 0.354 (0.093) & 0.890 (0.098) & \textbf{0.922 (0.066)} \\
				&  & macro-$ \text{F}_{1} $ & 0.583 (0.142) & 0.931 (0.054) & 0.344 (0.064) & 0.913 (0.089) & \textbf{0.952 (0.042)} \\
				\cline{2-8}
				& Hard Distribution & Accuracy & 0.987 (0.016) & \textbf{0.991 (0.007)} & 0.986 (0.010) & 0.988 (0.010) & 0.984 (0.012) \\
				&  & NMI & 0.951 (0.051) & \textbf{0.961 (0.030)} & 0.945 (0.037) & 0.951 (0.038) & 0.936 (0.043) \\
				&  & ARI & 0.961 (0.049) & \textbf{0.973 (0.022)} & 0.959 (0.029) & 0.965 (0.028) & 0.953 (0.034) \\
				&  & macro-$ \text{F}_{1} $ & 0.987 (0.016) & \textbf{0.991 (0.007)} & 0.986 (0.010) & 0.988 (0.010) & 0.984 (0.012) \\
				\cline{2-8}
				& Jain & Accuracy & 0.914 (0.056) & \textbf{1.000 (0.000)} & 0.851 (0.046) & 0.992 (0.026) & \textbf{1.000 (0.000)} \\
				&  & NMI & 0.605 (0.160) & \textbf{1.000 (0.000)} & 0.329 (0.127) & 0.964 (0.116) & \textbf{1.000 (0.000)} \\
				&  & ARI & 0.673 (0.177) & \textbf{1.000 (0.000)} & 0.451 (0.130) & 0.969 (0.102) & \textbf{1.000 (0.000)} \\
				&  & macro-$ \text{F}_{1} $ & 0.888 (0.067) & \textbf{1.000 (0.000)} & 0.796 (0.061) & 0.990 (0.035) & \textbf{1.000 (0.000)} \\
				\cline{2-8}
				& Pathbased & Accuracy & 0.715 (0.109) & 0.945 (0.049) & 0.528 (0.093) & 0.895 (0.081) & \textbf{0.965 (0.048)} \\
				&  & NMI & 0.510 (0.115) & 0.844 (0.120) & 0.152 (0.089) & 0.765 (0.155) & \textbf{0.904 (0.119)} \\
				&  & ARI & 0.422 (0.178) & 0.837 (0.129) & 0.097 (0.095) & 0.718 (0.202) & \textbf{0.895 (0.134)} \\
				&  & macro-$ \text{F}_{1} $ & 0.668 (0.105) & 0.943 (0.056) & 0.511 (0.086) & 0.894 (0.082) & \textbf{0.965 (0.050)} \\
				\hline
				Real-world & Breast Cancer & Accuracy & 0.897 (0.044) & \textbf{0.923 (0.031)} & 0.789 (0.272) & 0.911 (0.040) & 0.913 (0.041) \\
				&  & NMI & 0.549 (0.139) & \textbf{0.625 (0.105)} & 0.420 (0.185) & 0.581 (0.144) & 0.596 (0.149) \\
				&  & ARI & 0.626 (0.140) & \textbf{0.710 (0.102)} & 0.506 (0.208) & 0.674 (0.133) & 0.680 (0.135) \\
				&  & macro-$ \text{F}_{1} $ & 0.885 (0.048) & \textbf{0.915 (0.033)} & 0.779 (0.270) & 0.903 (0.044) & 0.905 (0.045) \\
				\cline{2-8}
				& COIL20 & Accuracy & 0.416 (0.051) & \textbf{0.832 (0.054)} & 0.797 (0.065) & 0.646 (0.076) & 0.646 (0.076) \\
				&  & NMI & 0.707 (0.042) & 0.887 (0.028) & \textbf{0.891 (0.032)} & 0.806 (0.037) & 0.806 (0.037) \\
				&  & ARI & 0.401 (0.054) & \textbf{0.756 (0.063)} & 0.741 (0.078) & 0.558 (0.065) & 0.558 (0.065) \\
				&  & macro-$ \text{F}_{1} $ & 0.305 (0.048) & \textbf{0.820 (0.057)} & 0.766 (0.080) & 0.579 (0.085) & 0.579 (0.085) \\
				\cline{2-8}
				& Iris & Accuracy & 0.907 (0.063) & \textbf{0.933 (0.048)} & \textbf{0.933 (0.061)} & 0.813 (0.146) & 0.913 (0.108) \\
				&  & NMI & 0.840 (0.076) & \textbf{0.864 (0.085)} & 0.860 (0.113) & 0.787 (0.131) & 0.861 (0.121) \\
				&  & ARI & 0.800 (0.107) & 0.830 (0.116) & \textbf{0.841 (0.137)} & 0.701 (0.196) & 0.826 (0.165) \\
				&  & macro-$ \text{F}_{1} $ & 0.868 (0.113) & \textbf{0.913 (0.074)} & 0.904 (0.100) & 0.759 (0.187) & 0.890 (0.137) \\
				\cline{2-8}
				& Isolet & Accuracy & 0.543 (0.072) & \textbf{0.779 (0.031)} & 0.535 (0.031) & 0.137 (0.034) & 0.137 (0.034) \\
				&  & NMI & 0.691 (0.036) & \textbf{0.803 (0.021)} & 0.683 (0.017) & 0.347 (0.040) & 0.347 (0.040) \\
				&  & ARI & 0.443 (0.056) & \textbf{0.646 (0.041)} & 0.421 (0.027) & 0.066 (0.022) & 0.066 (0.022) \\
				&  & macro-$ \text{F}_{1} $ & 0.504 (0.083) & \textbf{0.771 (0.033)} & 0.494 (0.046) & 0.053 (0.023) & 0.053 (0.023) \\
				\cline{2-8}
				& OptDigits & Accuracy & 0.899 (0.018) & \textbf{0.955 (0.009)} & 0.794 (0.038) & 0.683 (0.221) & 0.681 (0.219) \\
				&  & NMI & 0.830 (0.025) & \textbf{0.914 (0.017)} & 0.701 (0.032) & 0.636 (0.171) & 0.633 (0.168) \\
				&  & ARI & 0.796 (0.034) & \textbf{0.904 (0.020)} & 0.587 (0.076) & 0.530 (0.231) & 0.526 (0.227) \\
				&  & macro-$ \text{F}_{1} $ & 0.899 (0.018) & \textbf{0.955 (0.009)} & 0.802 (0.034) & 0.633 (0.270) & 0.631 (0.269) \\
				\cline{2-8}
				& Seeds & Accuracy & 0.867 (0.067) & 0.871 (0.054) & 0.869 (0.072) & 0.826 (0.147) & \textbf{0.888 (0.070)} \\
				&  & NMI & 0.707 (0.119) & 0.699 (0.105) & 0.692 (0.153) & 0.680 (0.154) & \textbf{0.736 (0.132)} \\
				&  & ARI & 0.647 (0.155) & 0.657 (0.119) & 0.661 (0.180) & 0.624 (0.195) & \textbf{0.699 (0.176)} \\
				&  & macro-$ \text{F}_{1} $ & 0.856 (0.082) & 0.858 (0.069) & 0.851 (0.074) & 0.809 (0.148) & \textbf{0.876 (0.075)} \\
				\cline{2-8}
				& Sonar & Accuracy & 0.486 (0.069) & 0.599 (0.091) & 0.650 (0.122) & \textbf{0.671 (0.144)} & \textbf{0.671 (0.144)} \\
				&  & NMI & 0.031 (0.042) & 0.066 (0.069) & 0.137 (0.203) & \textbf{0.241 (0.140)} & \textbf{0.241 (0.140)} \\
				&  & ARI & 0.005 (0.015) & 0.043 (0.065) & 0.118 (0.209) & \textbf{0.164 (0.179)} & \textbf{0.164 (0.179)} \\
				&  & macro-$ \text{F}_{1} $ & 0.402 (0.098) & 0.582 (0.099) & 0.631 (0.135) & \textbf{0.635 (0.157)} & \textbf{0.635 (0.157)} \\
				\cline{2-8}
				& TOX171 & Accuracy & 0.336 (0.098) & 0.406 (0.103) & N/A & \textbf{0.508 (0.124)} & \textbf{0.508 (0.124)} \\
				&  & NMI & 0.218 (0.120) & 0.290 (0.098) & N/A & \textbf{0.439 (0.155)} & \textbf{0.439 (0.155)} \\
				&  & ARI & 0.061 (0.073) & 0.066 (0.087) & N/A & \textbf{0.223 (0.132)} & \textbf{0.223 (0.132)} \\
				&  & macro-$ \text{F}_{1} $ & 0.247 (0.090) & 0.356 (0.122) & N/A & \textbf{0.409 (0.155)} & \textbf{0.409 (0.155)} \\
				\cline{2-8}
				& Wine & Accuracy & 0.637 (0.096) & 0.652 (0.099) & \textbf{0.866 (0.113)} & 0.777 (0.148) & 0.767 (0.188) \\
				&  & NMI & 0.460 (0.139) & 0.484 (0.111) & \textbf{0.711 (0.173)} & 0.604 (0.181) & 0.604 (0.223) \\
				&  & ARI & 0.328 (0.160) & 0.345 (0.146) & \textbf{0.632 (0.212)} & 0.486 (0.237) & 0.502 (0.278) \\
				&  & macro-$ \text{F}_{1} $ & 0.570 (0.092) & 0.624 (0.092) & \textbf{0.870 (0.114)} & 0.763 (0.164) & 0.752 (0.191) \\
				\hline\hline
			\end{tabular}
		}
	}
	%	\vspace{-4mm}
	\\
	\vspace{1mm}
	\footnotesize \raggedright \hspace{13mm}The best value in each metric is indicated by bold. The standard deviation is indicated in parentheses.
	
	\hspace{13mm}N/A indicates that an algorithm could not build a predictive model.
\end{table*}

Table \ref{tab:NodeNS} shows the number of generated nodes by each algorithm in the non-stationary environment. The general trend is the same as in the stationary environment, where GH-EXIN has a large number of generated nodes and GHNG has a small number of them. Regarding HCAEA, this algorithm generates a single layer for COIL20, Isolet, Sonar, TOX171, and Wine datasets, which are the same as in the stationary environment.

Table \ref{tab:TimeNS} summarizes the results of training time in the non-stationary environment. The general trend of the computation speed is also the same as in the stationary environment. Namely, GHNG is faster, the computation speed of GH-EXIN depends on the number of attributes, and the computation speed of HCAEA and CAEA depends on the number of generated nodes.

\subsection{Parameter Sensitivity}
\label{sec:paramSensitiveity}

\begin{figure*}[htbp]
	\centering
	%	\hfil
	\hspace{-1.4mm}
	\subfloat[Aggregation]{
		\includegraphics[width=1.35in]{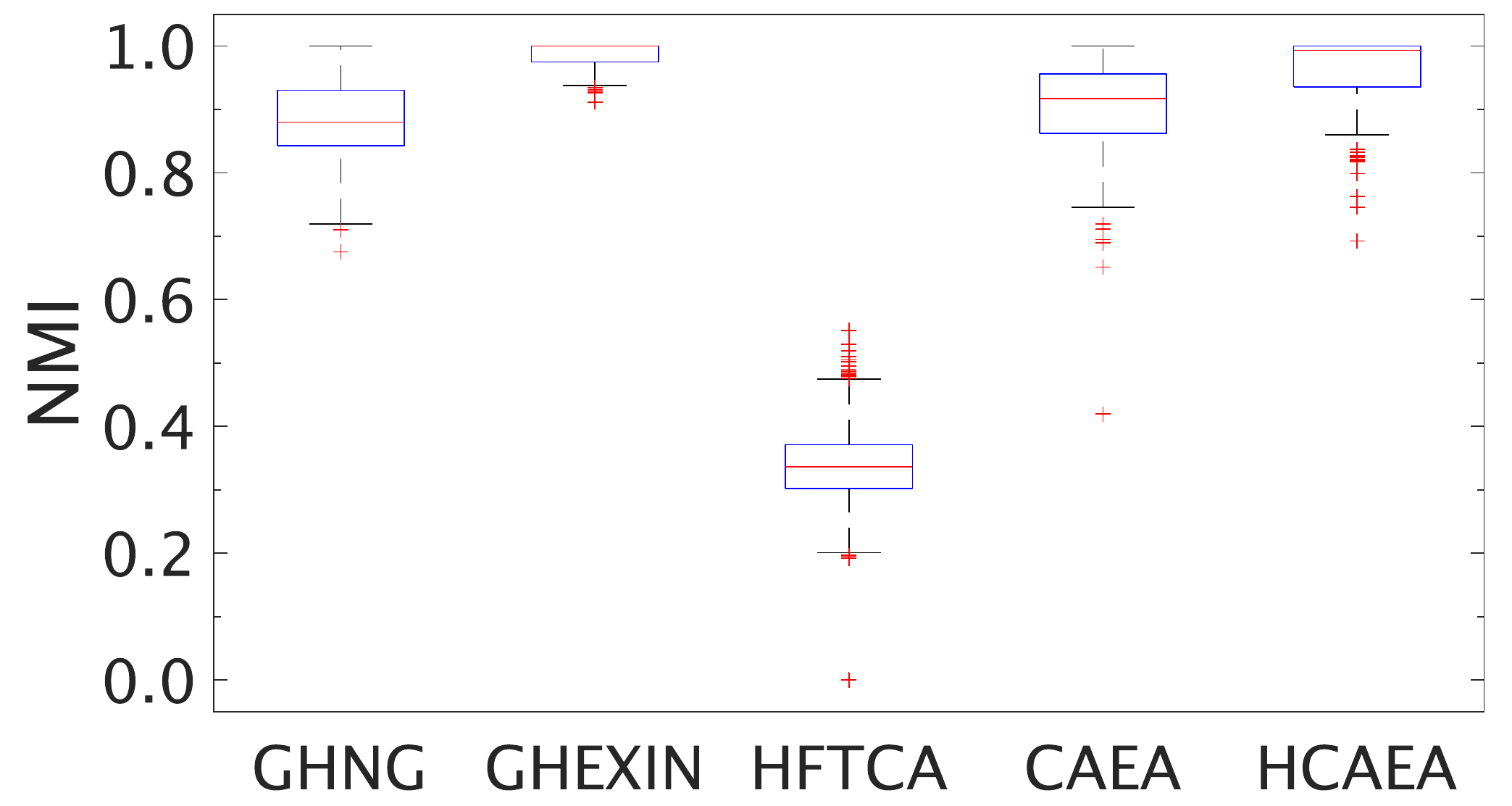}
		\label{fig:BP_Aggregation_S}
	}
	\hspace{-1.4mm}
	\hfil
	\subfloat[Compound]{
		\includegraphics[width=1.35in]{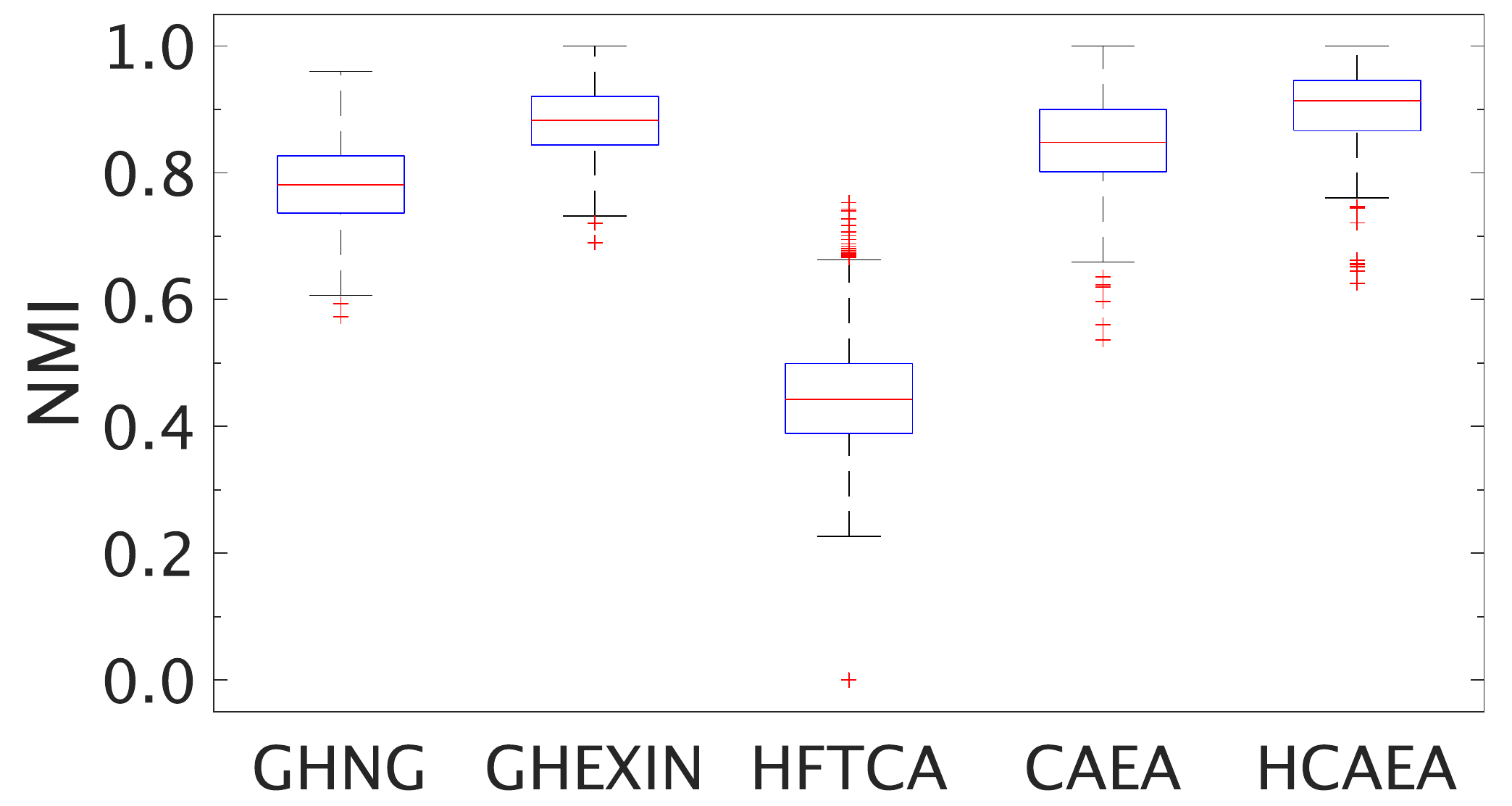}
		\label{fig:BP_Compound_S}
	}
	\hspace{-1.4mm}
	\hfil
	\subfloat[Hard Distribution]{
		\includegraphics[width=1.35in]{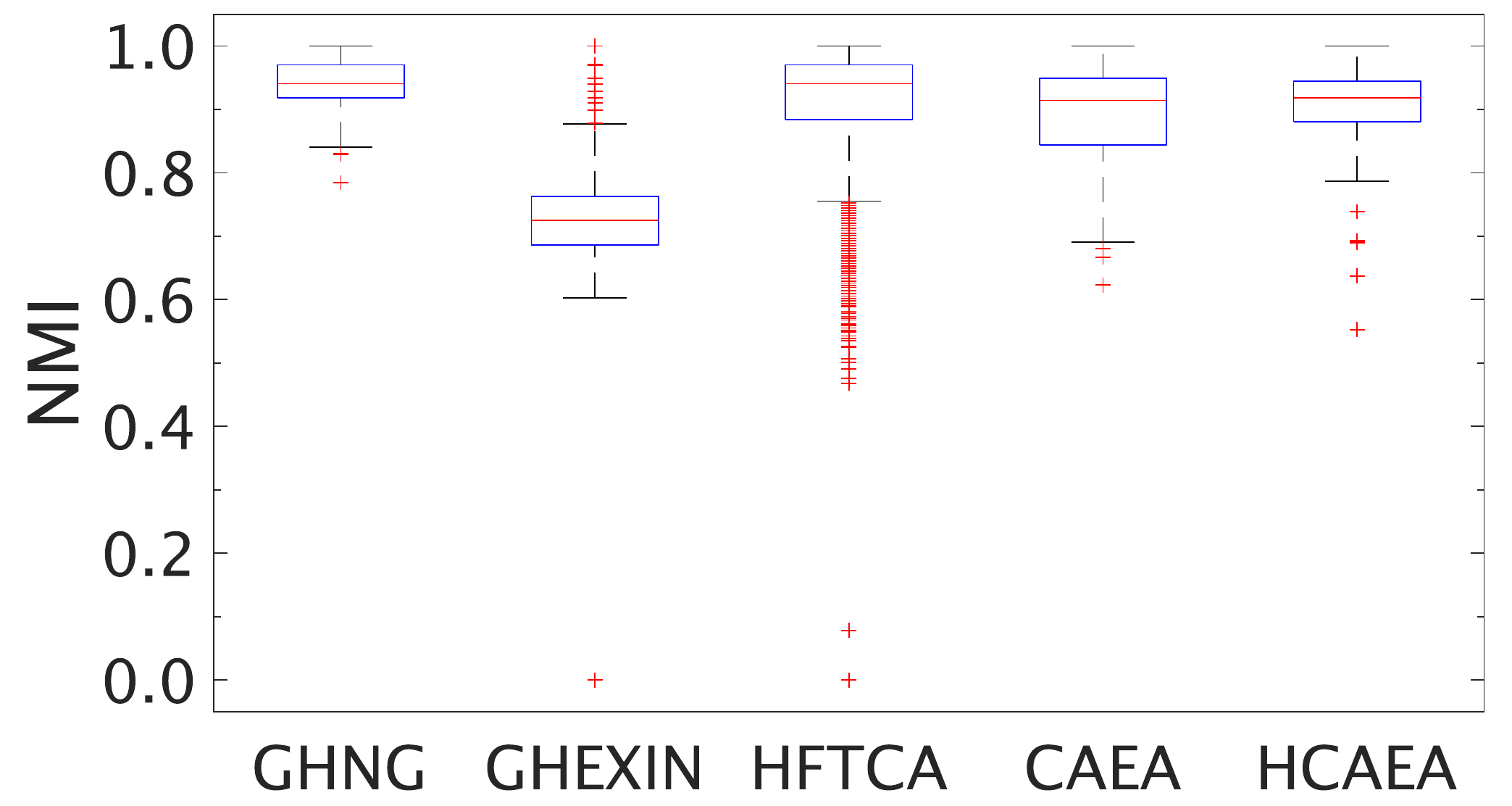}
		\label{fig:BP_HardDistribution_S}
	}
	\hspace{-1.4mm}
	\hfil
	\subfloat[Jain]{
		\includegraphics[width=1.35in]{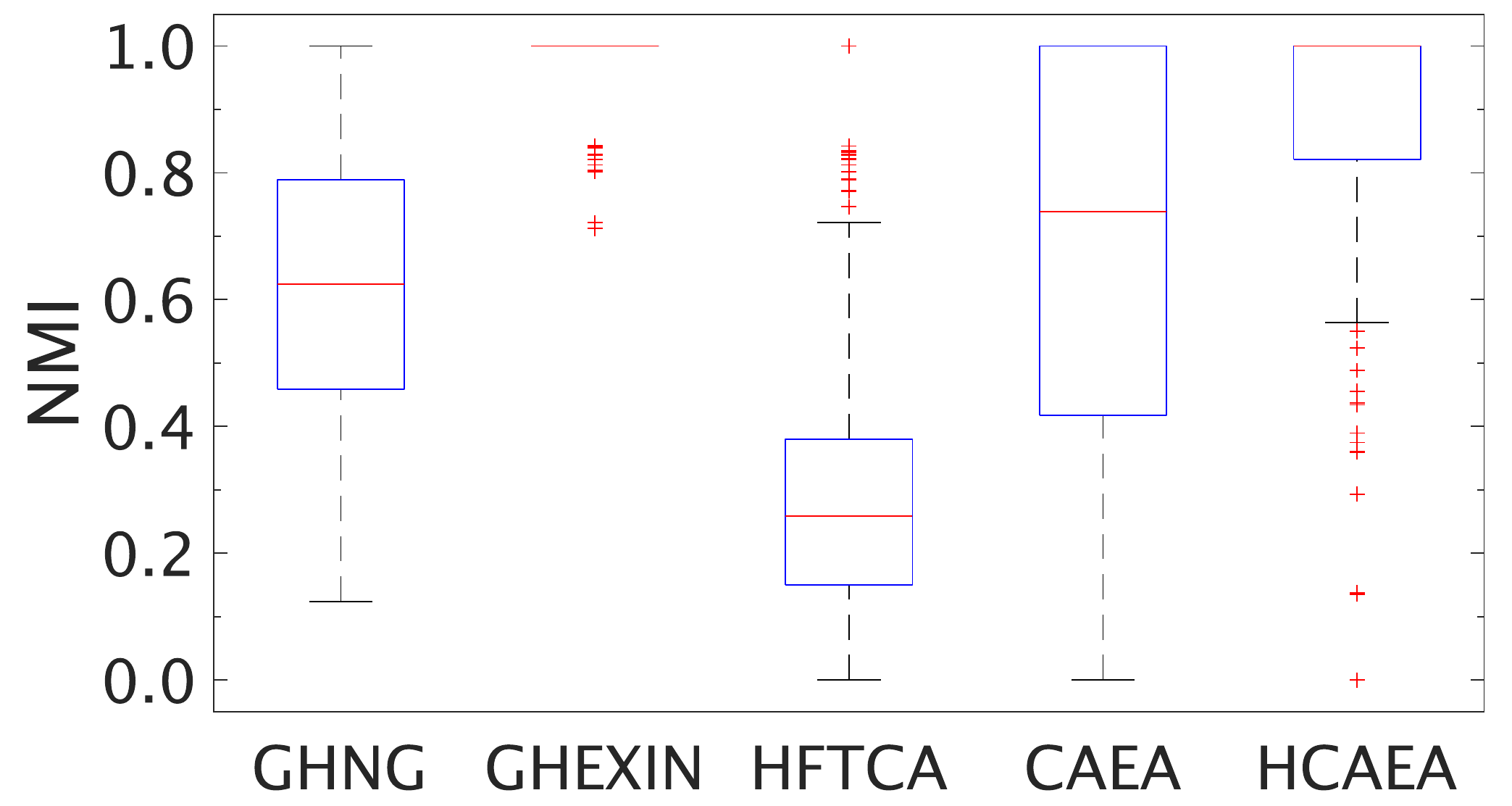}
		\label{fig:BP_Jain_S}
	}
	\hspace{-1.4mm}
	\hfil
	\subfloat[Pathbased]{
		\includegraphics[width=1.35in]{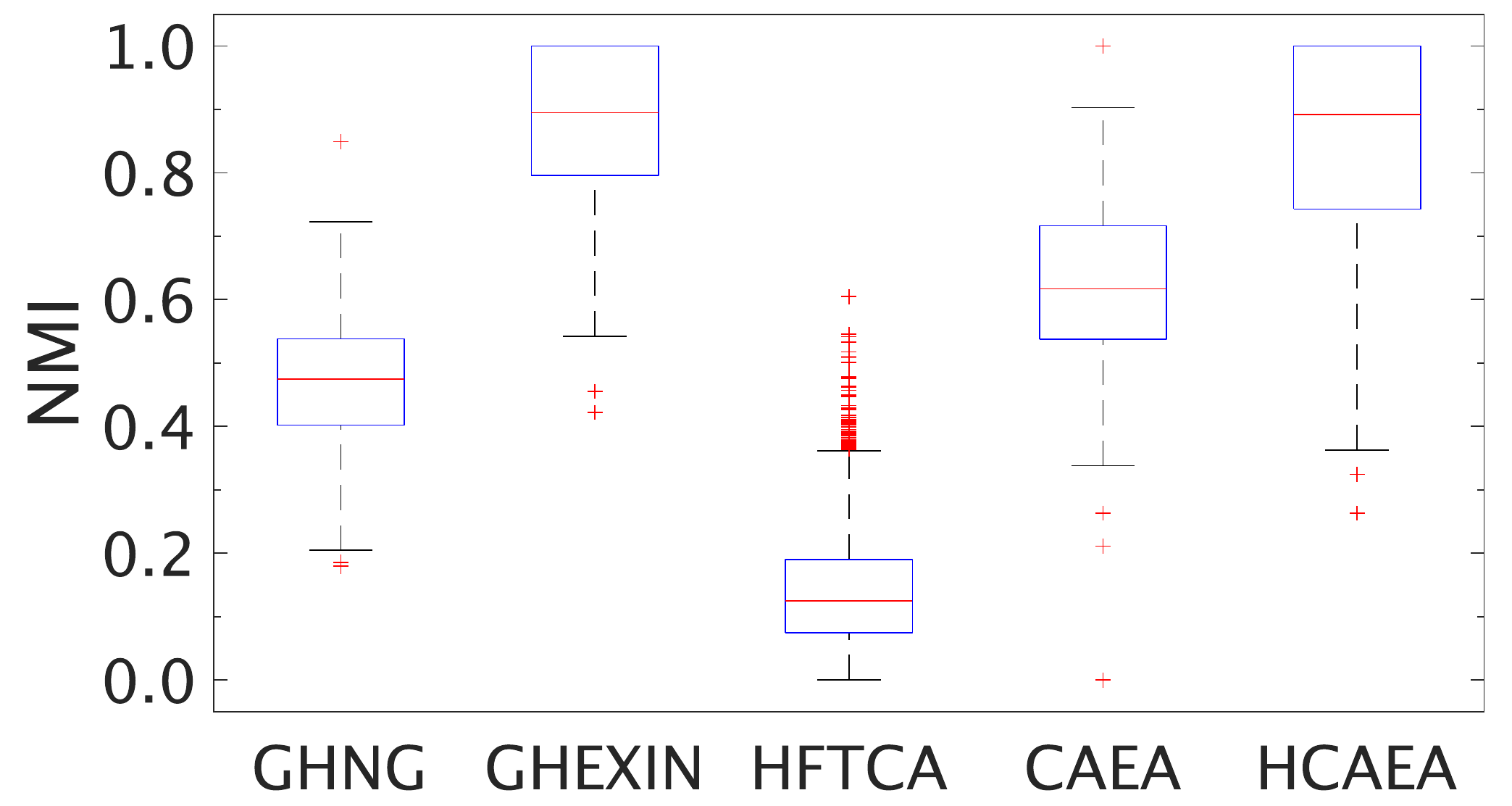}
		\label{fig:BP_Pathbased_S}
	}
	\\
	\hspace{-1.4mm}
	%	\hfil
	\subfloat[Breast Cancer]{
		\includegraphics[width=1.35in]{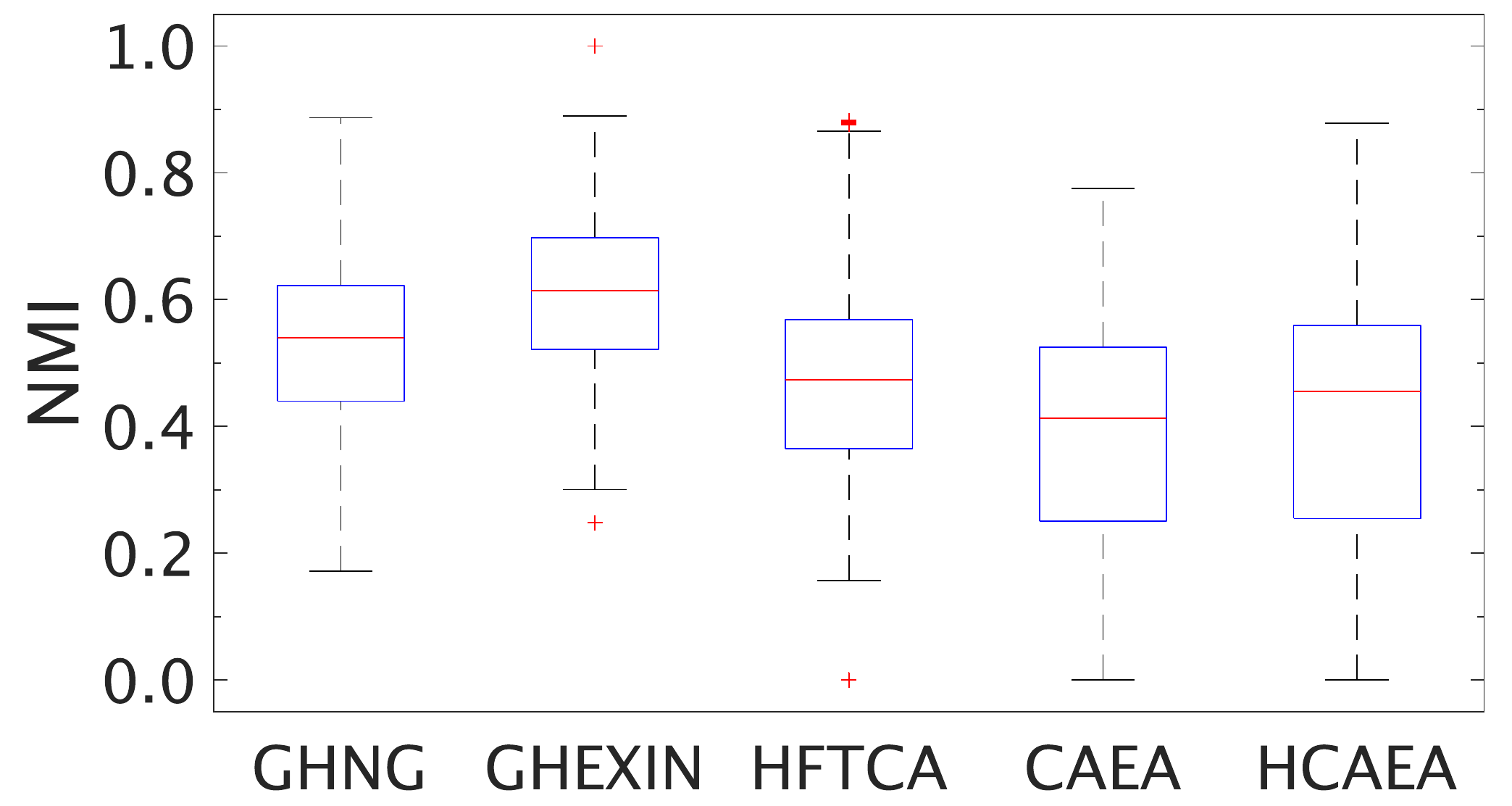}
		\label{fig:BP_BreastCancer_S}
	}
	\hspace{-1.4mm}
	\hfil
	\subfloat[COIL20]{
		\includegraphics[width=1.35in]{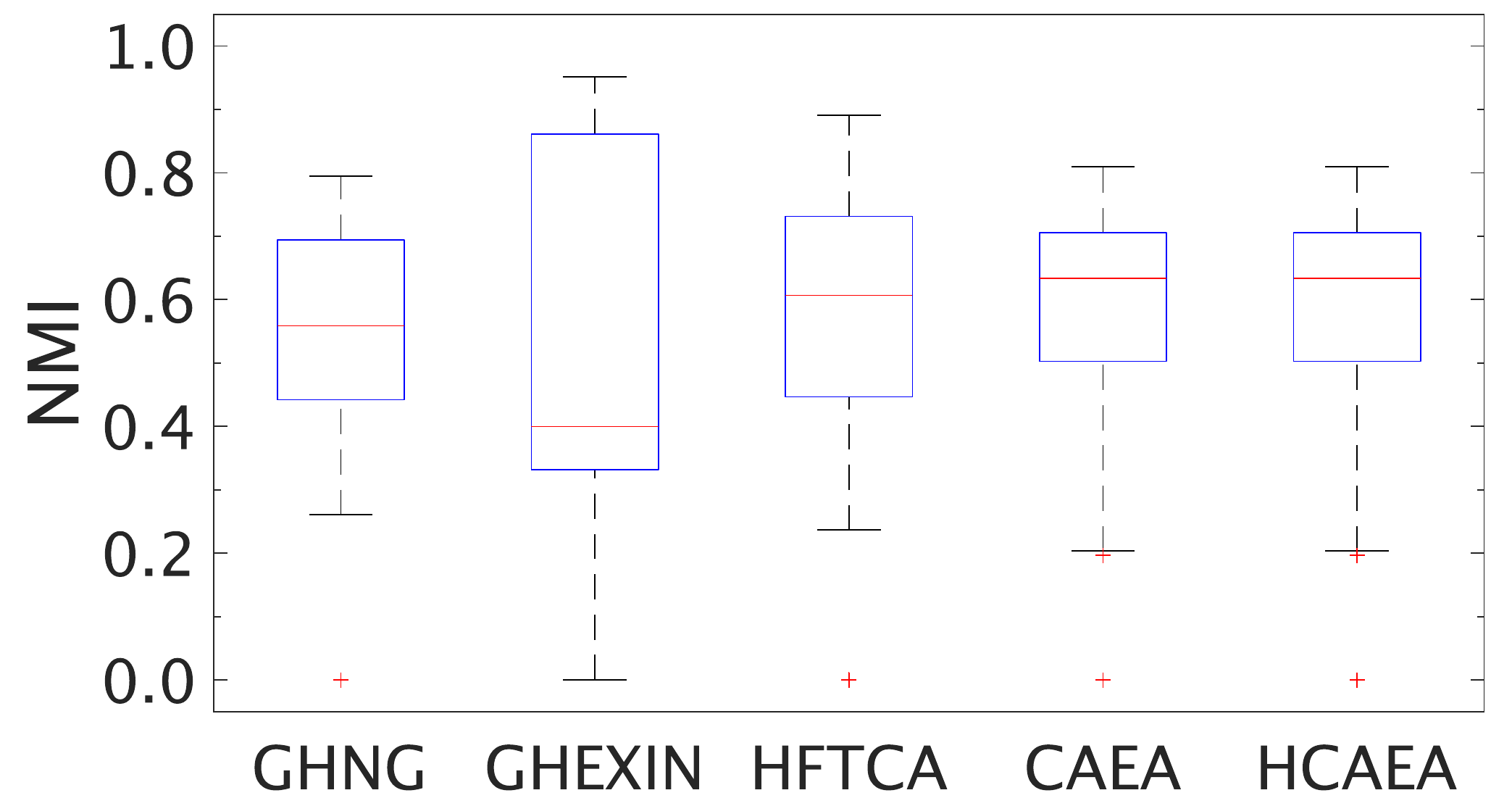}
		\label{fig:BP_COIL20_S}
	}
	\hspace{-1.4mm}
	\hfil
	\subfloat[Iris]{
		\includegraphics[width=1.35in]{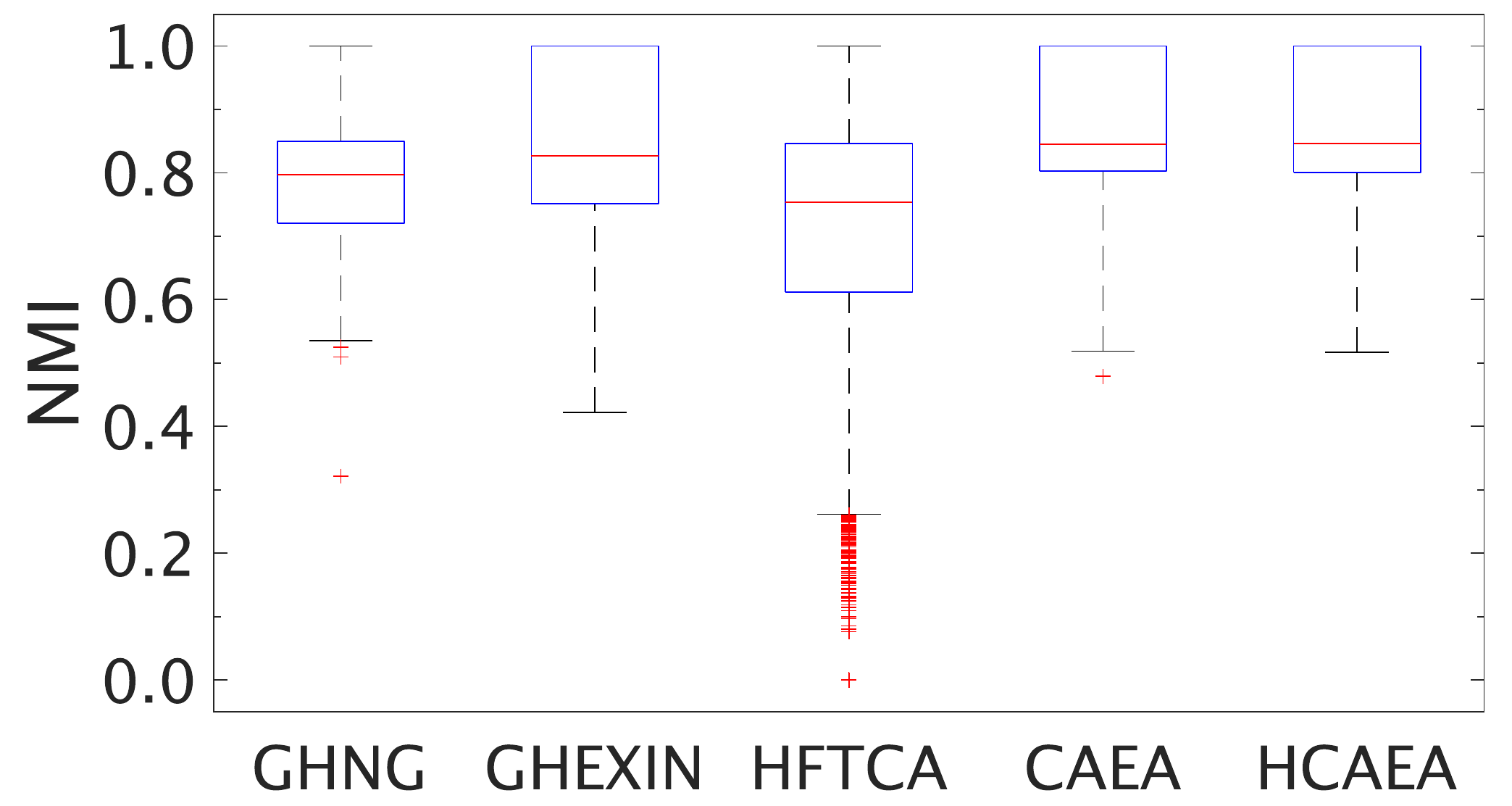}
		\label{fig:BP_Iris_S}
	}
	\hspace{-1.4mm}
	\hfil
	\subfloat[Isolet]{
		\includegraphics[width=1.35in]{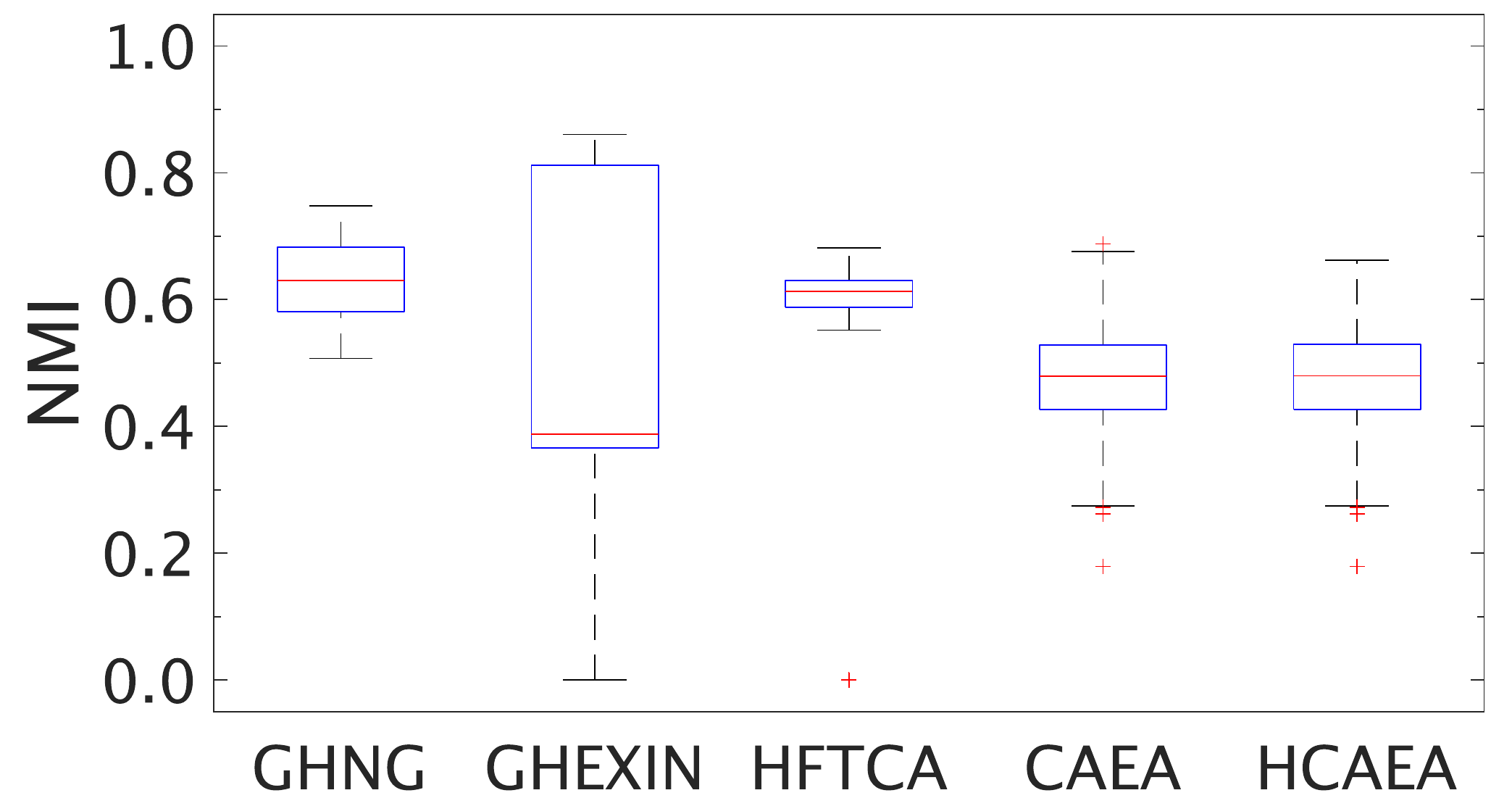}
		\label{fig:BP_Isolet_S}
	}
	\hspace{-1.4mm}
	\hfil
	\subfloat[OptDigits]{
		\includegraphics[width=1.35in]{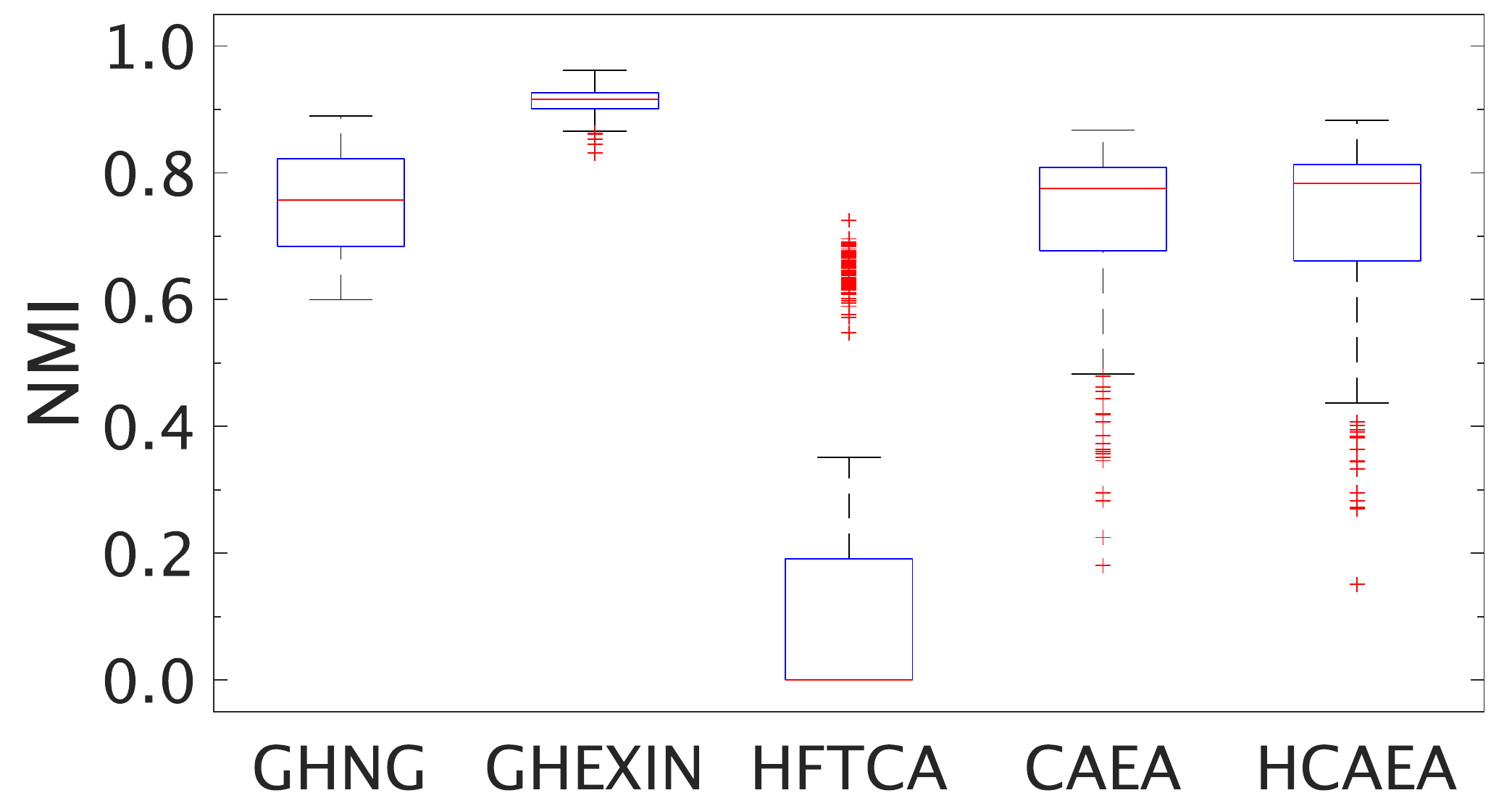}
		\label{fig:BP_OptDigits_S}
	}
	\\
	\vspace{0.0mm}
	%	\hfil
	\subfloat[Seeds]{
		\includegraphics[width=1.35in]{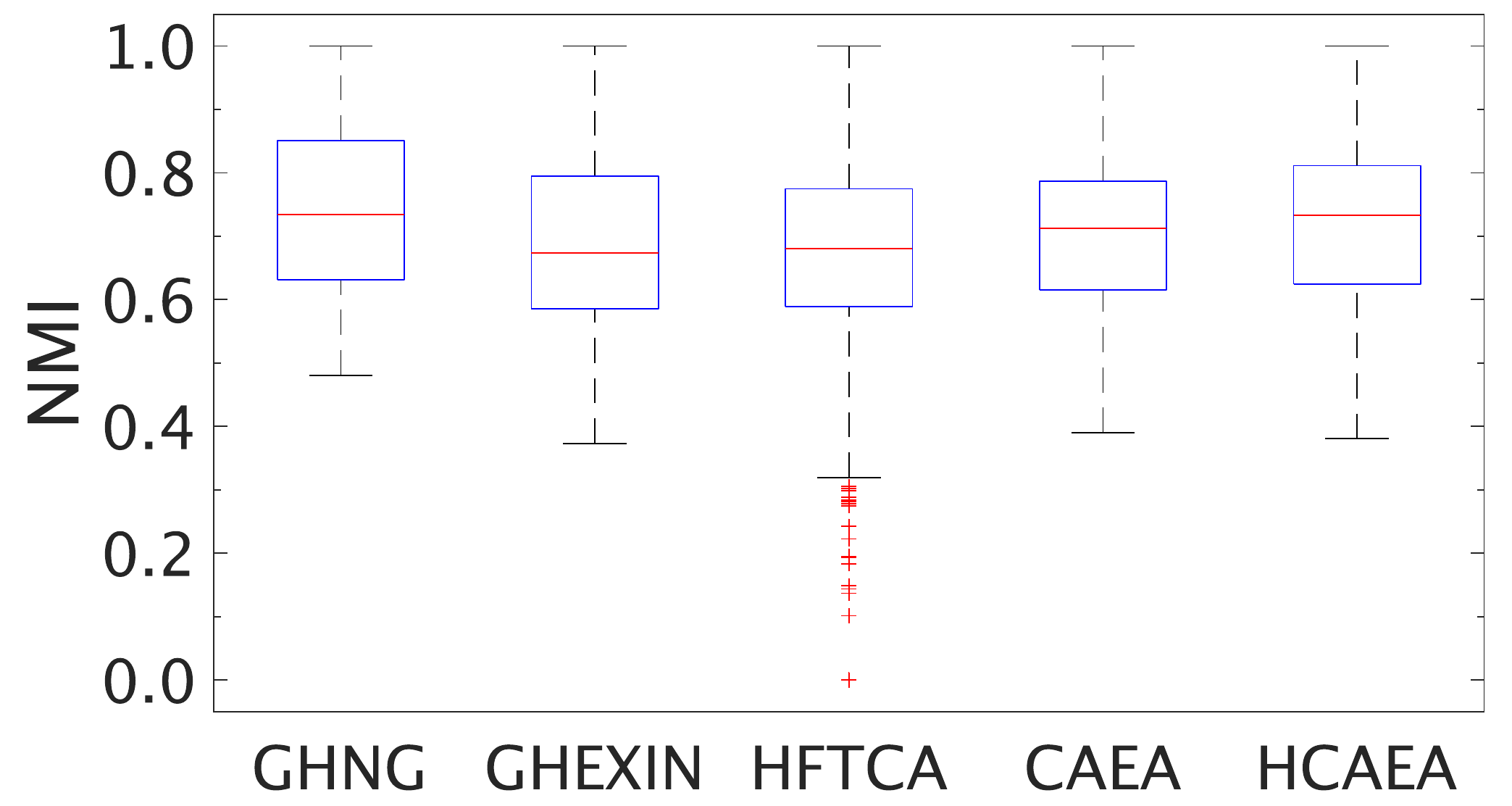}
		\label{fig:BP_Seeds_S}
	}
	\hspace{-1.4mm}
	%	\hfil
	\subfloat[Sonar]{
		\includegraphics[width=1.35in]{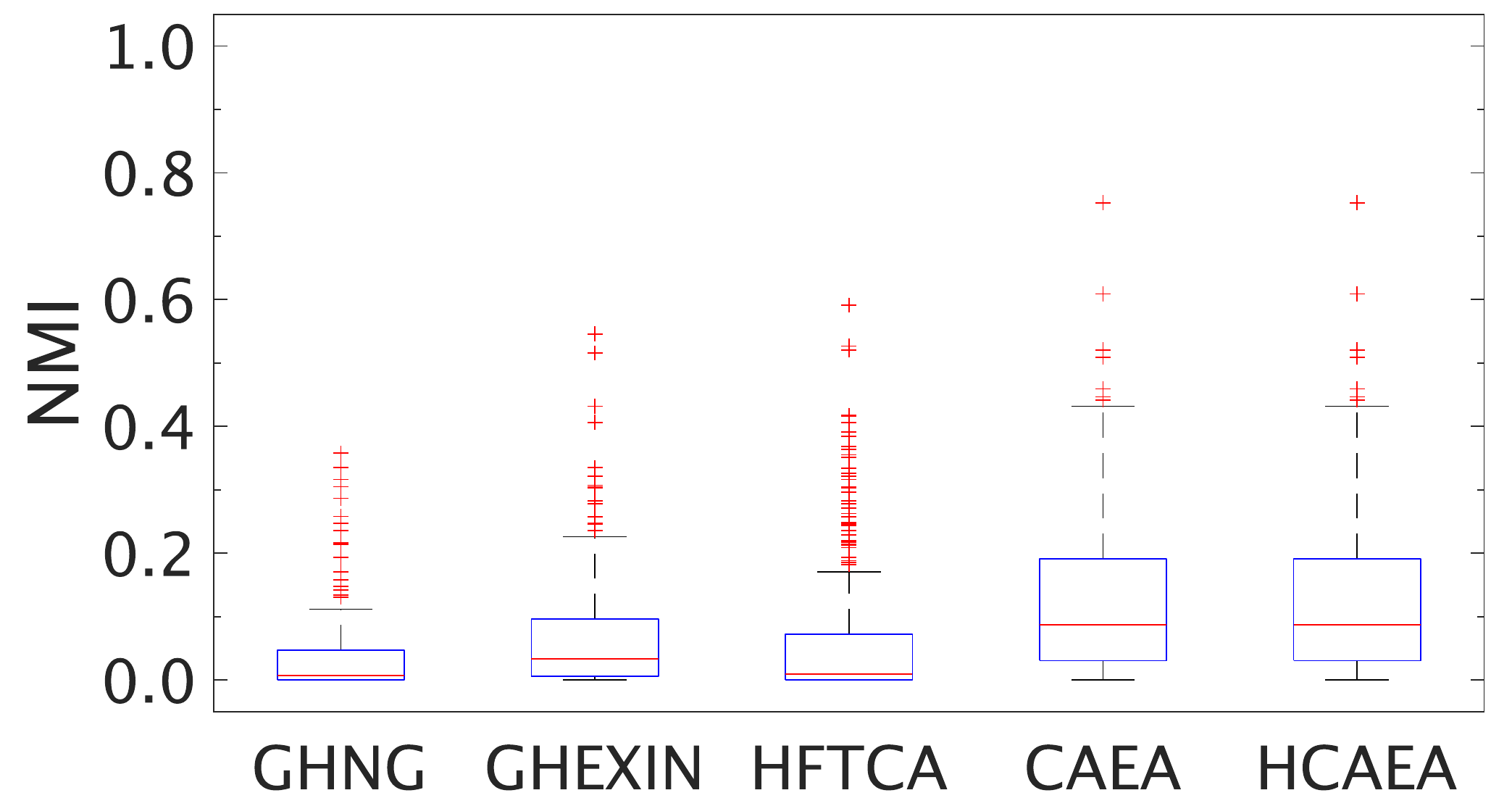}
		\label{fig:BP_Sonar_S}
	}
	\hspace{-1.4mm}
	%	\hfil
	\subfloat[TOX171]{
		\includegraphics[width=1.35in]{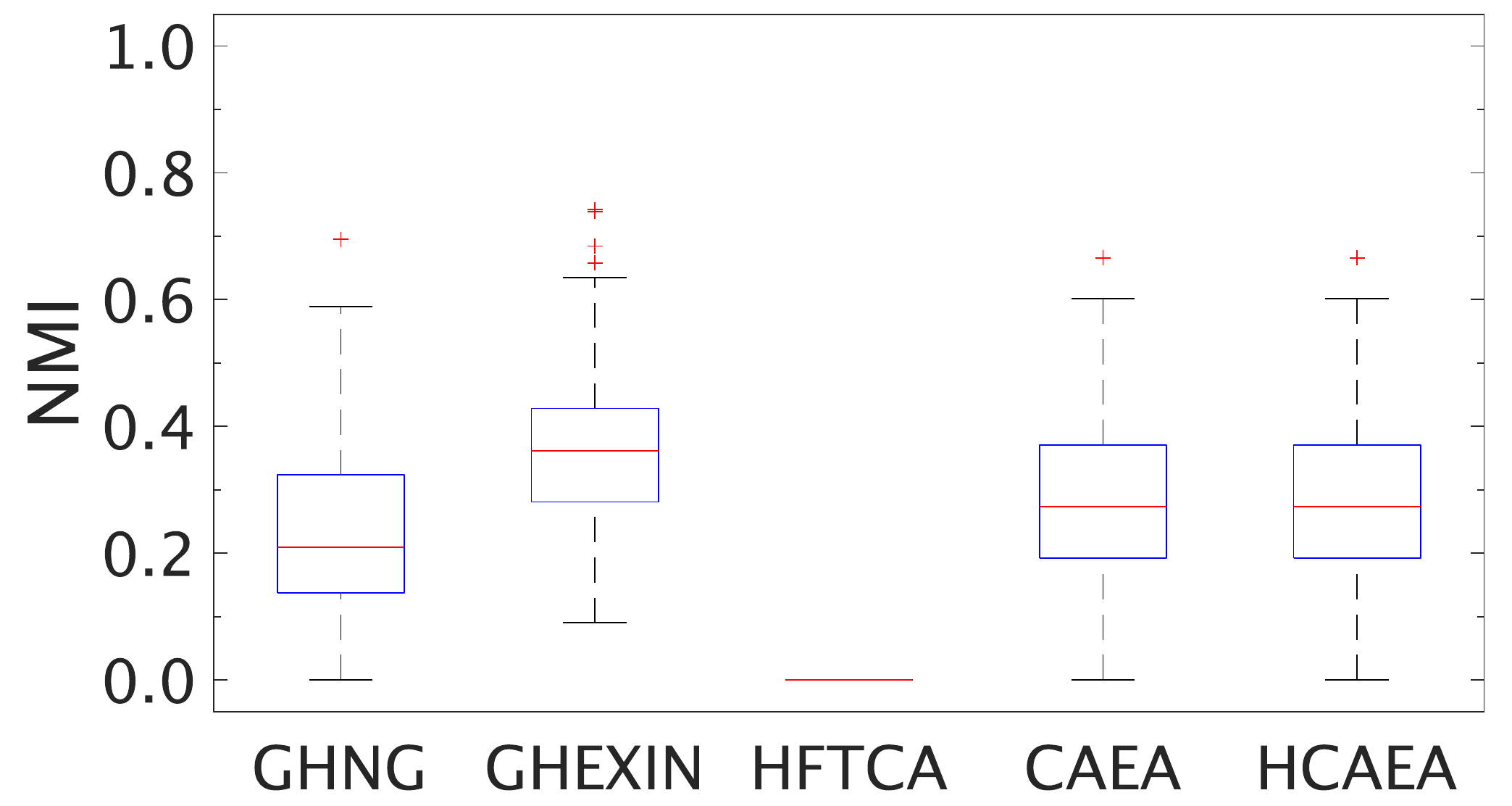}
		\label{fig:BP_TOX171_S}
	}
	\hspace{-1.4mm}
	%	\hfil
	\subfloat[Wine]{
		\includegraphics[width=1.35in]{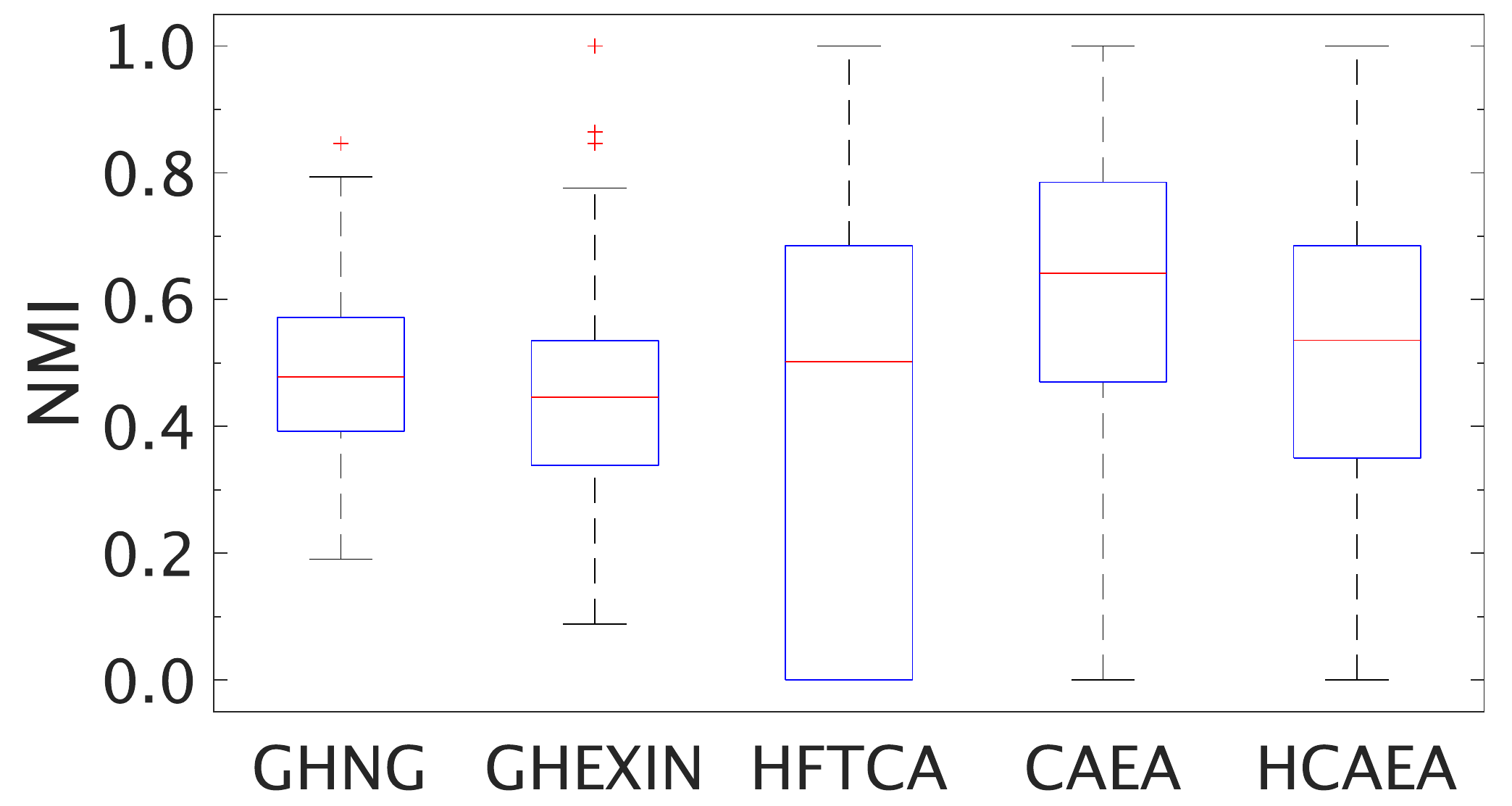}
		\label{fig:BP_Wine}
	}
	\caption{Box plot with NMI obtained by grid search in the stationary environment.}
	\label{fig:paramSensitivity_S}
\end{figure*}

Fig. \ref{fig:paramSensitivity_S} shows box plots with all the results of NMI obtained by grid search in Section \ref{sec:palamSpec} for each dataset in the stationary environment. Depending on the values of the parameters in the grid range, each algorithm shows high/low NMI. By observing the variations of NMI, the sensitivity of an algorithm to each parameter specification can be evaluated.

Although the median of NMI is different, as a general tendency, the interquartile ranges are similar for each algorithm on each dataset except for small datasets (e.g., Jain and Wine) and high-dimensional datasets (e.g., COIL20 and Isolet). In regard to the parameters specified by grid search (see Table \ref{tab:paramClassificationGrid}), those values for GHNG, GH-EXIN, and HFTCA are selected from the entire grid range. In contrast, the parameters of CAEA and HCAEA are mostly selected from a certain range, i.e., $ \{24, 26, 28, 30\} $, although the candidate of grid values are $ \{10, 12,\ldots, 30\} $. This observation suggests that CAEA and HCAEA can apply the same parameter setting to a wide variety of datasets.

\begin{table*}[htbp]
	\centering
	\renewcommand{\arraystretch}{1.1}
	\caption{
		Results of the number of leaf nodes in the non-stationary environment
	}
	\label{tab:NodeNS}
	\scalebox{1.0}{ 
		\begin{tabular}{ll|c|c|c|c|c}
			\hline\hline
			Type & Dataset & GHNG & GH-EXIN & HFTCA & CAEA & HCAEA \\
			\hline
			Synthetic & Aggregation & 23.9 (2.3) & 139.1 (8.7) & 107.1 (9.9) & 74.6 (6.5) & 498.0 (75.2) \\
			& Compound & 14.5 (1.6) & 130.1 (11.3) & 63.0 (5.5) & 29.9 (4.1) & 50.6 (24.0) \\
			& Hard Distribution & 14.0 (0.0) & 292.9 (34.1) & 71.7 (14.9) & 87.3 (9.3) & 87.3 (9.3) \\
			& Jain & 11.6 (1.6) & 93.4 (7.8) & 58.4 (6.6) & 39.0 (3.9) & 278.6 (32.5) \\
			& Pathbased & 39.5 (2.3) & 305.6 (11.6) & 46.4 (5.1) & 66.8 (9.6) & 559.9 (130.1) \\
			\hline
			Real-world & Breast Cancer & 3.0 (0.0) & 33.6 (3.3) & 48.4 (17.4) & 27.4 (3.4) & 40.4 (4.8) \\
			& COIL20 & 71.3 (12.4) & 1758.5 (238.6) & 66.7 (4.8) & 39.8 (2.3) & 39.8 (2.3) \\
			& Iris & 10.1 (1.4) & 87.0 (4.5) & 25.5 (3.7) & 38.8 (4.4) & 231.5 (35.2) \\
			& Isolet & 155.2 (8.1) & 1265.2 (97.7) & 56.3 (11.0) & 100.8 (51.5) & 110.8 (65.3) \\
			& OptDigits & 9.4 (1.2) & 74.5 (6.0) & 1091.4 (13.1) & 35.9 (3.5) & 191.6 (22.2) \\
			& Seeds & 3.5 (1.1) & 48.1 (3.9) & 43.5 (5.3) & 27.9 (3.0) & 50.8 (23.6) \\
			& Sonar & 3.5 (1.1) & 46.7 (5.9) & 16.8 (2.6) & 23.1 (2.0) & 23.1 (2.0) \\
			& TOX171 & 3.0 (0.0) & 33.2 (8.4) & N/A & 27.5 (3.6) & 27.5 (3.6) \\
			& Wine & 3.5 (1.1) & 42.8 (3.1) & 31.7 (3.3) & 24.6 (3.7) & 27.4 (4.0) \\
			\hline\hline
		\end{tabular}
	}
	\\
	\vspace{1mm}
	\footnotesize \raggedright \hspace{22mm}The standard deviation is indicated in parentheses.
	
	\hspace{22mm}N/A indicates that an algorithm could not build a predictive model.
\end{table*}

\begin{table*}[htbp]
	\centering
	\renewcommand{\arraystretch}{1.1}
	\caption{
		Results of training time on CPU [sec] in the non-stationary environment
	}
	\label{tab:TimeNS}
	\scalebox{1.0}{ 
		\begin{tabular}{ll|c|c|c|c|c}
			\hline\hline
			Type & Dataset & GHNG & GH-EXIN & HFTCA & CAEA & HCAEA \\
			\hline
			Synthetic & Aggregation & 43.125 (7.952) & 2.097 (0.320) & 110.278 (13.641) & 64.900 (2.274) & 76.470 (6.296) \\
			& Compound & 0.759 (0.066) & 2.638 (0.206) & 3.633 (0.276) & 0.817 (0.157) & 0.958 (0.241) \\
			& Hard Distribution & 0.158 (0.021) & 330.892 (151.078) & 46.797 (8.806) & 2.529 (0.229) & 2.573 (0.242) \\
			& Jain & 3.096 (0.293) & 0.918 (0.322) & 3.304 (0.237) & 4.443 (0.501) & 4.470 (0.931) \\
			& Pathbased & 63.483 (3.684) & 6.247 (0.645) & 2.199 (0.172) & 36.101 (2.311) & 60.177 (7.249) \\
			\hline
			Real-world & Breast Cancer & 0.244 (0.018) & 0.276 (0.025) & 0.759 (0.268) & 0.270 (0.037) & 0.301 (0.040) \\
			& COIL20 & 1.665 (0.134) & 379.113 (43.146) & 1.992 (0.182) & 12.973 (0.422) & 13.036 (0.629) \\
			& Iris & 2.961 (0.399) & 0.784 (0.073) & 0.316 (0.036) & 2.299 (0.718) & 4.003 (0.648) \\
			& Isolet & 36.363 (1.707) & 119.191 (10.419) & 7.894 (0.587) & 57.970 (10.203) & 60.253 (12.377) \\
			& OptDigits & 2.220 (0.182) & 0.791 (0.263) & 347.531 (15.500) & 1.355 (0.294) & 2.687 (0.396) \\
			& Seeds & 0.273 (0.030) & 0.428 (0.039) & 0.300 (0.021) & 0.257 (0.086) & 0.377 (0.162) \\
			& Sonar & 0.029 (0.003) & 0.964 (0.114) & 0.042 (0.002) & 0.058 (0.010) & 0.056 (0.007) \\
			& TOX171 & 0.002 (0.000) & 204.276 (29.950) & N/A & 0.023 (0.002) & 0.024 (0.002) \\
			& Wine & 0.105 (0.014) & 0.383 (0.033) & 0.110 (0.013) & 0.101 (0.026) & 0.104 (0.026) \\
			\hline\hline
		\end{tabular}
	}
	\\
	\vspace{1mm}
	\footnotesize \raggedright \hspace{12mm}The standard deviation is indicated in parentheses.
	
	\hspace{12mm}N/A indicates that an algorithm could not build a predictive model.
\end{table*}

\subsection{Computational Complexity}
\label{sec:complexity}
This section presents the computational complexity of CAEA, HCAEA, and comparison algorithms. Specifically, CAEA and HCAEA are analyzed in detail.

Here, assuming $ n $ is the number of data points, $ \lambda $ is an interval for adapting $ \sigma $, $ d $ is a dimension of a data point and a node, and $ K $ is the number of generated nodes. In CAEA, the computational complexity of each process is as follows: Computing a bandwidth of a kernel function in the CIM is $ \mathcal{O}(\frac{n}{\lambda}d) $ (line 5 in Alg. \ref{al:trainCAEA}). Computing the CIM is $ \mathcal{O}(nd\frac{\lambda}{2}) $ (line 7 in Alg. \ref{al:trainCAEA}), and for sorting the result of the CIM is $ \mathcal{O}(\frac{\lambda}{2}\log{\frac{\lambda}{2}}) $ (line 7 in Alg. \ref{al:trainCAEA}). Similarly, for computing the CIM in line 9 is $ \mathcal{O}(ndK) $ and for sorting the result of the CIM in line 9 is $ \mathcal{O}(K\log{K}) $. In general, $ \lambda \ll n $ and $ K \ll n $. Thus, the total computational complexity of CAEA is $ \mathcal{O}( nd(\lambda + K) ) $.

In HCAEA, assuming that the number of data points in the $ m_{l} $th partition of the $ l $th layer of HCAEA is $ n_{lm_{l}} $, then the computational complexity is $ \mathcal{O}( \frac{n_{lm_{l}}}{\lambda}d + n_{lm_{l}}d\frac{\lambda}{2} + \frac{\lambda}{2}\log{\frac{\lambda}{2}} + n_{lm_{l}}dK_{lm_{l}}  + K_{lm_{l}}\log{K_{lm_{l}}} ) $. Here, $ K_{lm_{l}} $ is the number of generated nodes in the $ m_{l} $th partition of the $ l $th layer. Thus, the total computational complexity from the 2nd to the $ L $th layer is $ \mathcal{O}( \sum_{l = 2}^{L} \sum_{i = 1}^{m_{l}} (\frac{n_{li}}{\lambda}d + n_{li}d\frac{\lambda}{2} + \frac{\lambda}{2}\log{\frac{\lambda}{2}} + n_{li}dK_{li}  + K_{li}\log{K_{li}}) ) $. In general, $ \lambda \ll n $ and $ K \ll n $. As a result, the total computational complexity of HCAEA is $ \mathcal{O}( nd(\lambda + K) + \sum_{l = 2}^{L} \sum_{i = 1}^{m_{l}} n_{li} d (\lambda + m_{l} )  ) $.

A self-organizing process of GHNG uses GNG but its hierarchical approach is similar to HCAEA. The computational complexity of GNG is $ \mathcal{O}( nK ) $. Thus, the total computational complexity from the 2nd to the $ L $th layer is $ \mathcal{O}( \sum_{l = 2}^{L} \sum_{i = 1}^{m_{l}} n_{li} K_{li} ) $. As a result, the total computational complexity of GHNG is $ \mathcal{O}( nK + \sum_{l = 2}^{L} \sum_{i = 1}^{m_{l}} n_{li} K_{li} ) $.

Regarding GH-EXIN, its computational complexity is analyzed as $ \mathcal{O}( bJn\log_{b}{n} ) $ \cite{cirrincione20} where $ b $ is the average branching factor, $ J $ is the average number of epochs, and $ n $ is the number of data points. Note that further information can be found in \cite{cirrincione20}.

A learning algorithm and a hierarchical approach of HFTCA is the same with HCAEA except for a calculation of the vigilance parameter. The computational complexity of the calculation of the vigilance parameter can be ignored here. Thus, the computational complexity of HFTCA is $ \mathcal{O}( nd(\lambda + K) + \sum_{l = 2}^{L} \sum_{i = 1}^{m_{l}} n_{li} d (\lambda + m_{l} )  ) $.

The computational complexity for each algorithm is summarized in Table \ref{tab:compComplexity}.

 \begin{table}[htbp]
	\vspace{2mm}
	\centering
	\caption{Summary of computational complexity}
	\renewcommand{\arraystretch}{1.5}
	\label{tab:compComplexity}
	%	\begin{tabular}{llcp{28.8em}} 
	\scalebox{1.0}{
		\begin{tabular}{ll}
			\hline\hline
			Algorithm & Computational Complexity  \\
			\hline
			GHNG  &  $ \mathcal{O}( nK + \sum_{l = 2}^{L} \sum_{i = 1}^{m_{l}} n_{li} K_{li} ) $ \\
			\hline
			GH-EXIN  & $ \mathcal{O}( bJn\log_{b}{n} ) $    \\
			\hline
			HFTCA  & $ \mathcal{O}( nd(\lambda + K) + \sum_{l = 2}^{L} \sum_{i = 1}^{m_{l}} n_{li} d (\lambda + m_{l} )  ) $ \\
			\hline
			CAEA  & $ \mathcal{O}( nd(\lambda + K) ) $ \\
			\hline
			HCAEA  & $ \mathcal{O}( nd(\lambda + K) + \sum_{l = 2}^{L} \sum_{i = 1}^{m_{l}} n_{li} d (\lambda + m_{l} )  ) $ \\
			\hline\hline
		\end{tabular}
	}
\end{table}

\section{Discussion}
\label{sec:discussion}
This section summarizes the characteristics of each algorithm based on the results in Sections \ref{sec:stationary} and \ref{sec:nonstationary} for emphasizing properties of HCAEA and CAEA.

In regard to GHNG, the computation speed is faster than all the other algorithms. However, the information extraction performance is inferior to the others. One possible reason for its poor performance is that a network generated by GNG does not adequately approximate the distribution of the data points in each dataset. This drawback can be resolved by increasing the number of training epochs, but this is a major disadvantage in terms of the concept of continual learning.

GH-EXIN maintains the superior information extraction performance both in the stationary and non-stationary environments. However, GH-EXIN tends to generate a large number of nodes, and the computation speed of GH-EXIN depends on the number of attributes in the dataset. Although GH-EXIN shows superior information extraction performance, the above characteristics become disadvantages when the algorithm is applied to a real-world dataset with a large number of attributes. The common disadvantage of the GNG-based algorithms (i.e., GHNG and GH-EXIN), is that there are a lot of parameters to be specified in advance (see Table \ref{tab:paramAlgorithms}). Moreover, the specified parameter values are significantly for different datasets (see Table \ref{tab:paramClassificationGrid}). 

Regarding HFTCA, its information extraction performance is similar to GHNG, and inferior to GH-EXIN, CAEA, and HCAEA. In addition, the vigilance parameter $ V $, which is a predefined parameter, has to be specified to each layer in advance. If the value of the vigilance parameter $ V $ is not properly defined, the algorithm cannot generate a sufficient number of nodes to approximate the distribution of data points. Specifically, as shown in Section IV-C, since TOX171 is a high-dimensional data with a small number of data points, HFTCA could not build the predictive model because generated nodes are deleted as isolated nodes before the network (i.e., edge connections) is fully constructed. This problem may be avoided by giving training data points multiple times until the network is sufficiently constructed.

CAEA maintains better information extraction performance than GHNG and HFTCA both in the stationary and non-stationary environments. Moreover, CAEA and HCAEA can apply the same parameter setting to a wide variety of datasets. One disadvantage of HCAEA and CAEA is that these algorithms tends to have poor classification performance over comparison algorithms on datasets with a large number of classes (i.e., COIL20, Isolet, and OptDigits). This is because that CAEA (and also HCAEA) estimates the similarity threshold $ V $ and a bandwidth $ \sigma $ for a kernel function from a very small number of training data points compared to whole data points, the similarity threshold $ V $ and the bandwidth $ \sigma $ are not suitable for the distribution of data points if the number of classes is large. This problem could be solved by adapting a large value of $ \lambda $. As shown in Figs. \ref{fig:CD_S} and \ref{fig:CD_NS}, a unique drawback of CAEA is that the information extraction performance is significantly different between the stationary and non-stationary environments. By introducing a hierarchical structure to CAEA, HCAEA successfully solves the disadvantage of CAEA.  In general, HCAEA shows better information extraction performance than CAEA and comparable performance to GH-EXIN without the specification of a large number of parameters. One problem of HCAEA is that the algorithm tends to generate excessive nodes in the case of Iris and Pathbased datasets. However, this problem can be solved by implementing a mechanism to delete unnecessary nodes in HCAEA.

\section{Conclusion}
\label{sec:conclusion}
This paper proposed an ART-based topological clustering algorithm, called CAEA. CAEA automatically estimates a similarity threshold, i.e., a vigilance parameter, from the distributions of data points. In addition, a divisive hierarchical clustering algorithm capable of continual learning, called HCAEA, was also proposed by applying a hierarchial structure to CAEA. The experimental results showed that the hierarchical structure of HCAEA improves the information extraction performance while solving the disadvantage of CAEA, i.e., the information extraction performance is significantly different between the stationary and non-stationary environments. Moreover, it has been shown that HCAEA has various advantages (i.e. a small number of parameters, an automated mechanism for hierarchical structure design) and comparable clustering performance to recently-proposed state-of-the-art hierarchical clustering algorithms.

In this paper, we focused only on the avoidance of catastrophic forgetting in order to achieve stable continual learning. However, dealing with concept drift, i.e., the change of the concepts in the learned information, is also important for maintaining a continual learning ability \cite{lu18}. Therefore, as a future research topic, we will consider dealing with concept drift in HCAEA in order to extend the functionality of the algorithm.

\bibliographystyle{IEEEtran}
\bibliography{myref}

\EOD

\end{document}